\documentclass{article}

\usepackage{microtype}
\usepackage{graphicx}
\usepackage{wrapfig}
\usepackage{subfigure}
\usepackage{booktabs} 
\usepackage{float} 
\usepackage{hyperref}
\usepackage{colortbl}  
\usepackage{xcolor}    
\usepackage{adjustbox} 
\usepackage{csquotes}
\definecolor{highlight}{gray}{0.9}

\usepackage[accepted]{icml2025}

\usepackage{amsmath}
\usepackage{amssymb}
\usepackage{mathtools}
\usepackage{amsthm}
\usepackage{caption}
\usepackage{multirow}
\makeatletter

\long\def\@makecaption#1#2{
 \vskip 10pt
        \baselineskip 11pt
        \setbox\@tempboxa\hbox{#1. #2}
        \ifdim \wd\@tempboxa >\hsize
        \sbox{\newcaptionbox}{\small\sl #1.~}
        \newcaptionboxwid=\wd\newcaptionbox
        \usebox\newcaptionbox {\footnotesize #2}
        \else
          \centerline{{\small\sl #1.} {\small #2}}
        \fi}
\makeatother
\usepackage[capitalize,noabbrev]{cleveref}

\theoremstyle{plain}

\theoremstyle{definition}

\theoremstyle{remark}

\usepackage[textsize=tiny]{todonotes}

\icmltitlerunning{Enhance-A-Video: Better Generated Video for Free}

\begin{document}

\twocolumn[
\icmltitle{Enhance-A-Video: Better Generated Video for Free}



\begin{icmlauthorlist}
\icmlauthor{Yang Luo$^{1}$}{}
\icmlauthor{Xuanlei Zhao$^{1}$}{}
\icmlauthor{Mengzhao Chen$^{2}$}{}
\icmlauthor{Kaipeng Zhang$^{2}$}{}
\\
\icmlauthor{Wenqi Shao$^{2\dagger}$}{}
\icmlauthor{Kai Wang$^{1\dagger}$}{}
\icmlauthor{Zhangyang Wang$^{3}$}{}
\icmlauthor{Yang You$^{1}$}{}
\end{icmlauthorlist}

\begin{center}
$^1$National University of Singapore 
\quad $^2$Shanghai Artificial Intelligence Laboratory \quad $^3$University of Texas at Austin\\
\end{center}

\vskip 0.15in

{\begin{center}
\label{fig:abstract}
    \captionsetup{type=figure}
    \includegraphics[width=0.85\textwidth]{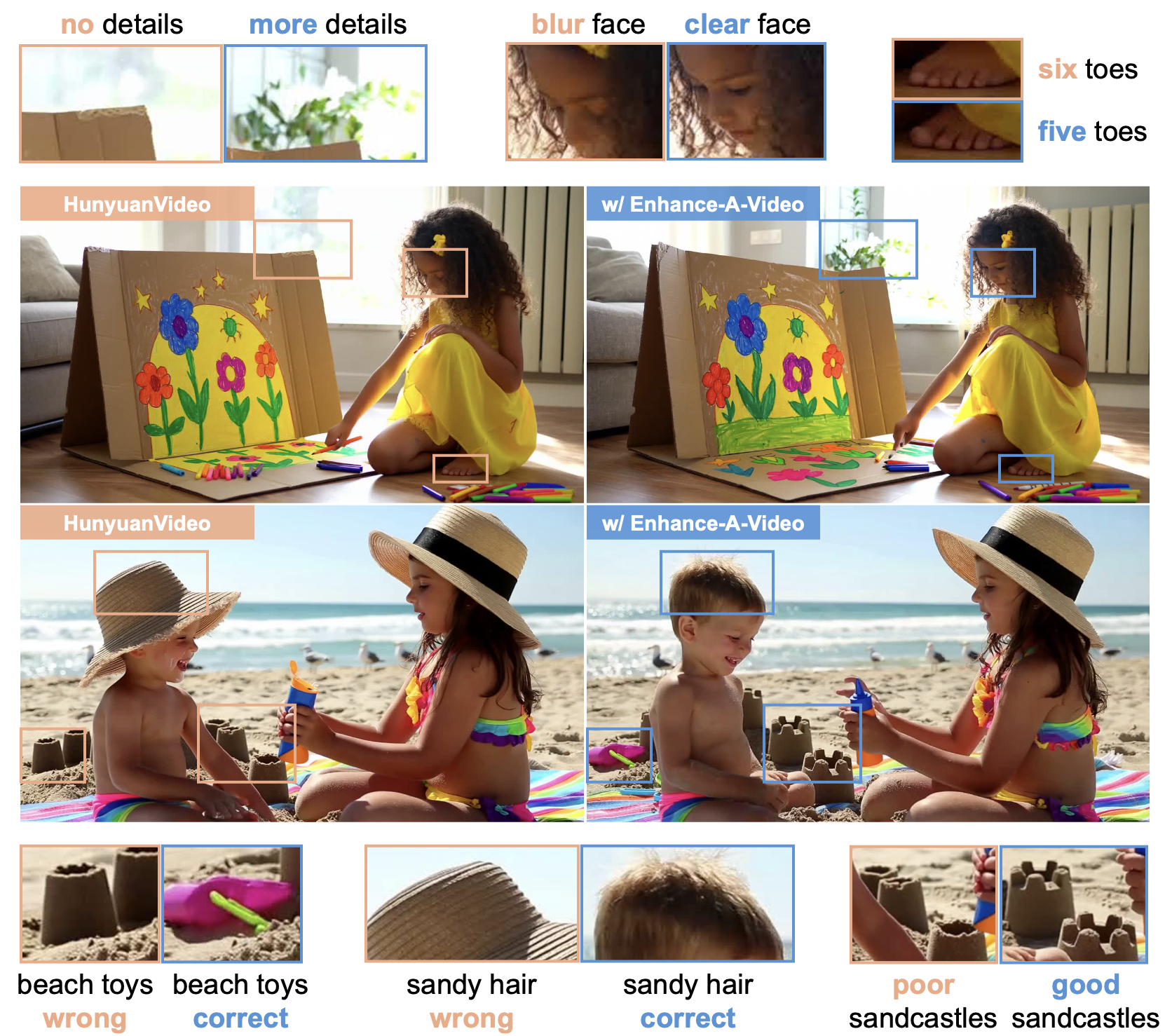}
    \vskip -0.1in
    \captionof{figure}{Enhance-A-Video boosts diffusion transformers-based video generation quality at minimal cost - no training needed, no extra learnable parameters, no memory overhead. Detailed captions are available in Appendix \ref{appendix:abstract}.}
\end{center}}

\vskip 0.1in
]
{
\renewcommand{\thefootnote}{}
\footnotetext[0]{$^{\dagger}$corresponding author}
\footnotetext[1]{Code: \href{https://github.com/NUS-HPC-AI-Lab/Enhance-A-Video}{NUS-HPC-AI-Lab/Enhance-A-Video}}
}

\begin{abstract}
DiT-based video generation has achieved remarkable results, but research into enhancing existing models remains relatively unexplored. In this work, we introduce a training-free approach to enhance the coherence and quality of DiT-based generated videos, named Enhance-A-Video. The core idea is enhancing the cross-frame correlations based on non-diagonal temporal attention distributions. Thanks to its simple design, our approach can be easily applied to most DiT-based video generation frameworks without any retraining or fine-tuning. Across various DiT-based video generation models, our approach demonstrates promising improvements in both temporal consistency and visual quality. We hope this research can inspire future explorations in video generation enhancement.
\end{abstract}

\section{Introduction}
\begin{figure}[H]
    \centering
    \includegraphics[width=0.95\linewidth]{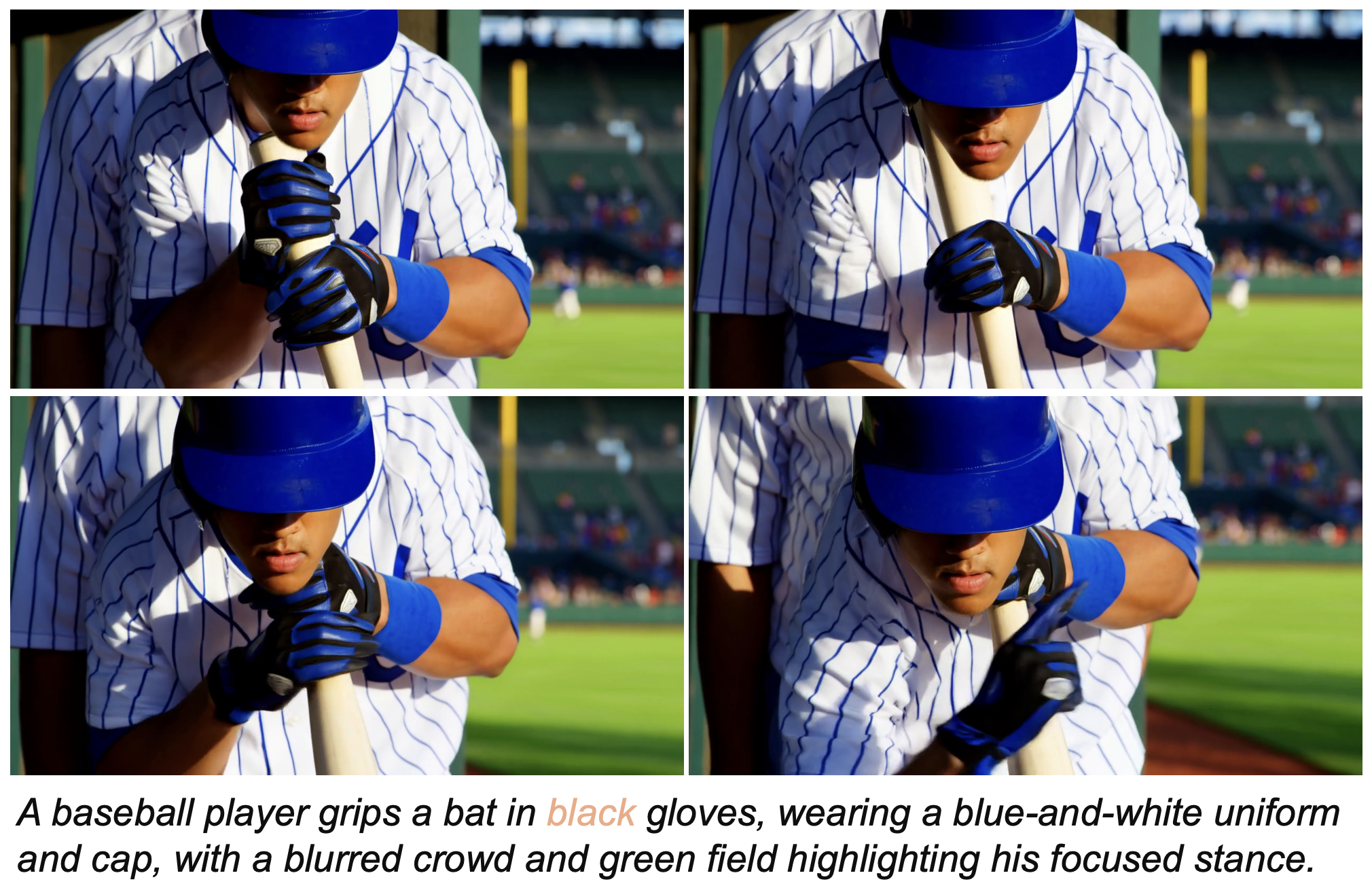} 
    \vskip -0.15in
    \caption{Video sample of HunyuanVideo model with \textit{unnatural head movements}, \textit{repeated right hands} and \textit{conflicting glove color.}}
    \label{fig:hunyuan_bad}
\vskip -0.15in
\end{figure}

Diffusion transformer (DiT) models \cite{DiT} have revolutionized video generation,  enabling the creation of realistic and compelling videos \cite{yang2024cogvideoxtexttovideodiffusionmodels, videoworldsimulators2024, lin2024opensoraplanopensourcelarge, xu2024easyanimatehighperformancelongvideo,
kong2025hunyuanvideosystematicframeworklarge}. However, achieving temporal consistency across frames while maintaining fine-grained details remains a significant challenge. Many existing methods generate videos that suffer from unnatural transitions and degraded quality as illustrated in Figure \ref{fig:hunyuan_bad}, which fundamentally limits their practical applicability in real-world scenarios and professional applications \cite{pmlr-v202-yan23b, henschel2024streamingt2v}. 

Video generation enhancement \cite{he2024venhancergenerativespacetimeenhancement} is designed for addressing the above limitations, where two objectives are primarily considered: (\textbf{$i$}) maintaining temporal consistency across frames, which ensures smooth and coherent transitions, and (\textbf{$ii$}) improving spatial details, which enhances the visual quality of each frame. In UNet-based 
video generation \cite{zhang2023show, guo2023animatediff, xu2024videogigagandetailrichvideosuperresolution, li2025diffvsrenhancingrealworldvideo}, Upscale-A-Video \cite{zhou2024upscaleavideo} integrated a local-global temporal strategy for better temporal coherence, and VEnhancer \cite{he2024venhancergenerativespacetimeenhancement} designed a video ControlNet \cite{zhang2023adding} to enhance spatial and temporal resolution simultaneously.
Nevertheless, the exploration of enhancing DiT-based video generation remains limited, particularly in addressing challenges of temporal consistency and spatial detail preservation.

\begin{figure*}[t]
    \centering
    \includegraphics[width=0.9\linewidth]{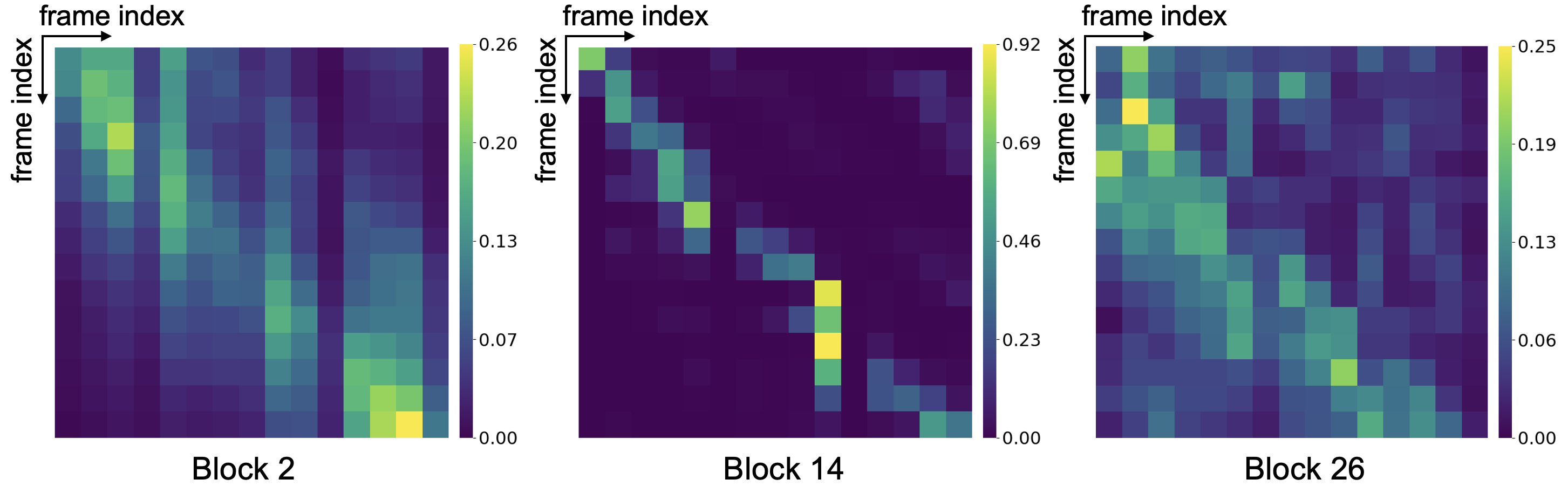} 
    \vskip -0.1in
    \caption{Visualization of temporal attention distributions in Open-Sora for blocks 2, 14, and 26 at denoising step 30, where non-diagonal elements are considerably weaker than diagonal elements.}
    \label{fig:attn_map}
\vskip -0.15in
\end{figure*}

In DiT-based video generation, temporal attention \cite{tan2023temporal} plays a crucial role in ensuring coherence among frames, further preserving fine-grained details.
Through careful analysis of temporal attention in DiT blocks, we made an important observation as shown in Figure \ref{fig:attn_map}: cross-frame temporal attentions (non-diagonal elements) are significantly lower than intra-frame attentions (diagonal elements) in some blocks. This unbalanced distribution of cross-frame and intra-frame attention may lead to inconsistencies among frames, such as abrupt transitions and blurred details in generated videos.

\textit{Is there an efficient method to utilize the cross-frame information to improve consistency across frames?} The intensity of cross-frame information is directly related to the mean of non-diagonal temporal attention weights.
By leveraging the calculated cross-frame intensity, 
it becomes possible to promote video quality by adjusting imbalanced cross-frame dependencies while maintaining frame-level detail.

Building on these insights,  we propose a novel, training-free, and plug-and-play approach, Enhance-A-Video, to improve the temporal and spatial quality of DiT-based generated videos. 
The method introduces two key innovations: a cross-frame intensity to capture cross-frame information within the temporal attention mechanism and an enhance temperature parameter to scale calculated cross-frame intensity. By strengthening cross-frame correlations from the temperature perspective, our approach enhances temporal consistency and preserves fine visual details effectively.
A notable advantage is that this method can be readily integrated into prevalent DiT-based video generation frameworks with negligible computational overhead.

We conduct a comprehensive experimental evaluation of our approach across several benchmark DiT-based video generation models including HunyuanVideo \cite{kong2025hunyuanvideosystematicframeworklarge}, CogVideoX \cite{yang2024cogvideoxtexttovideodiffusionmodels}, LTX-Video \cite{HaCohen2024LTXVideo} and Open-Sora \cite{zheng2024opensorademocratizingefficientvideo}. By incorporating Enhance-A-Video during the inference phase, these models demonstrate a significant improvement in generated video quality by reducing temporal inconsistencies and refining visual fidelity with minimal extra cost. 

\section{Related Work}
\textbf{Video Generation.} Recent advancements in video generation have been driven by powerful diffusion transformer-based models \cite{chen2024gentrondiffusiontransformersimage, ma2024lattelatentdiffusiontransformer, gao2024luminat2xtransformingtextmodality, lu2024freelongtrainingfreelongvideo}. Sora \cite{videoworldsimulators2024} has demonstrated exceptional capabilities in generating realistic and long-duration videos, establishing itself as a significant milestone in text-to-video generation. CogVideoX \cite{yang2024cogvideoxtexttovideodiffusionmodels} introduced a 3D full attention mechanism and expert transformers to improve motion consistency and semantic alignment. HunyuanVideo \cite{kong2025hunyuanvideosystematicframeworklarge} introduces a hybrid stream block with enhanced semantic understanding. 
However, several important challenges such as temporal inconsistency and the loss of fine-grained spatial details in video generation still persist.

\textbf{Temperature Parameter.} The temperature parameter is a well-known concept in deep learning, primarily used to control the distribution of attention or output probabilities in generative models \cite{peeperkorn2024temperature, Renze2024TheEO}. In natural language generation tasks, the temperature is often adjusted during inference to modulate the diversity of the generated text \cite{holtzman2020curious}. 
A higher temperature increases randomness, promoting creativity, while a lower temperature encourages deterministic and coherent outputs. 
Recently, the concept has been explored in vision-related tasks, such as visual question answering and multimodal learning \cite{chen2021empirical}, where temperature adjustments are applied to balance multimodal attention distributions. However, its application in DiT-based video generation, particularly in enhancing temporal attention, remains underexplored. 

\section{Methodology}
\subsection{Diffusion Transformer Models}

Diffusion Transformer models are inspired by the success of diffusion models in generating high-quality images and videos by iteratively refining noisy data \cite{ho2022video, blattmann2023stable, esser2024scaling}. These models combine the strengths of diffusion processes and transformer architectures to model temporal and spatial dependencies in video generation. The forward diffusion process adds noise to the data over $T$ timesteps, gradually converting it into a noise distribution. Starting from clean data $\mathbf{x}_0$, the noisy data at timestep $t$ is obtained as:
\begin{equation}
\mathbf{x}_t = \sqrt{\alpha_t} \mathbf{x}_{t-1} + \sqrt{1 - \alpha_t} \mathbf{z}_t, \quad \text{for } t = 1, \dots, T,
\end{equation}
where $\alpha_t$ controls the noise schedule and $\mathbf{z}_t \sim \mathcal{N}(0, \mathbf{I})$ is Gaussian noise. As $t$ increases, $\mathbf{x}_t$ approaches a standard normal distribution $\mathcal{N}(0, \mathbf{I})$. 
To recover the original data distribution, the reverse diffusion process progressively removes noise from $\mathbf{x}_t$ until reaching $\mathbf{x}_0$:
\begin{equation}
p_\theta(\mathbf{x}_{t-1} | \mathbf{x}_t) = \mathcal{N}(\mathbf{x}_{t-1}; \mu_\theta(\mathbf{x}_t, t), \Sigma_\theta(\mathbf{x}_t, t)),
\end{equation}
where $\mu_\theta$ and $\Sigma_\theta$ are learned parameters representing the mean and covariance of the denoised distribution.

\subsection{Temporal Attention in DiT Blocks}
\begin{figure*}[t]
    \centering
    \includegraphics[width=0.75\linewidth]{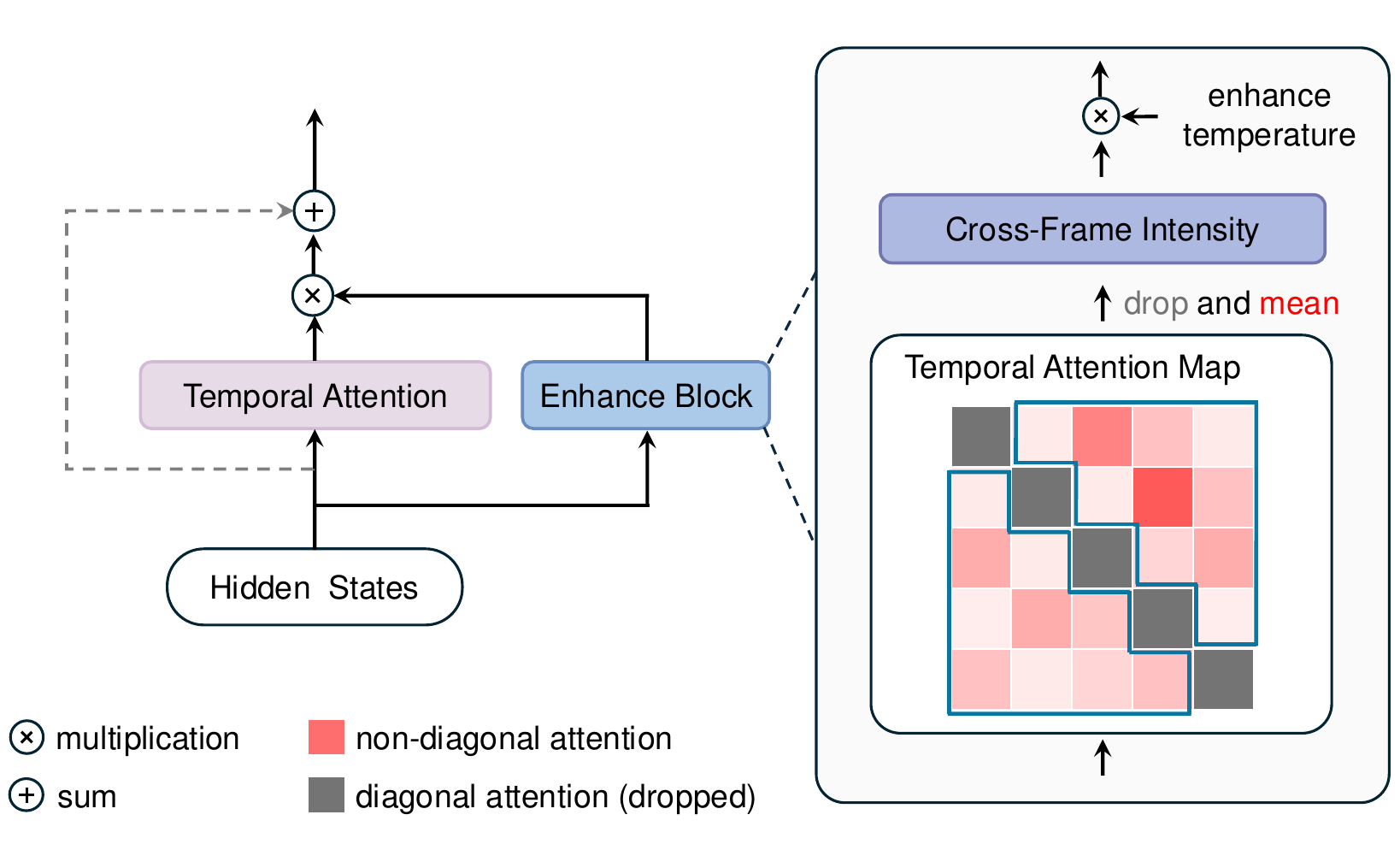} 
    \caption{Overview of the Enhance Block. The block computes the average of non-diagonal elements from the temporal attention map as Cross-Frame Intensity (\textit{CFI}). The \textit{CFI} is scaled by the temperature parameter and fused back to enhance the temporal attention output.}
    \label{fig:enhance_block}
\vskip -0.15in
\end{figure*}

DiT-based video generation models employ temporal transformer blocks focusing on cross-frame interactions. Each temporal block consists of feed-forward networks, self-attention modules, and optional cross-attention layers.

The temporal self-attention module computes attention weights between frames, allowing the model to aggregate information from past and future frames.
For video latent $\mathbf{z} \in \mathbb{R}^{B \times F \times C \times H \times W}$ with batch size $B$, $F$ frames, $C$ channels, and spatial dimensions $H \times W$, it reshapes features by merging spatial dimensions into the batch size, yielding $\tilde{\mathbf{z}} \in \mathbb{R}^{(B \times H \times W) \times F \times C}$. Self-attention \cite{vaswani2023attentionneed} is then applied along the frame axis:
\begin{equation}
A = \text{Attention}(Q(\tilde{\mathbf{z}}), K(\tilde{\mathbf{z}})) \in \mathbb{R}^{(B \times H \times W) \times F \times F}
\end{equation}
where $Q$ and $K$ denote the Query and Key heads, and $A$ satisfies $\sum_{j=1}^{F} A_{(b,i,j)} = 1$.

Temporal attention is crucial for balancing coherence and flexibility in video generation. However, findings in Figure \ref{fig:attn_map} reveal that standard temporal attention mechanisms often underemphasize cross-frame interactions, as attention weights for non-diagonal elements are typically much lower than diagonal elements. This shortcoming can lead to temporal inconsistencies like flickering or unexpected transitions, further affecting the spatial content negatively.

\subsection{Temperature in DiT-based Video Generation}
The temperature is a critical concept in large language model (LLM) inference, controlling the randomness and coherence of the generated tokens. The probability $P(x)$ of generating a token $x$ is adjusted using the temperature $\tau$ as:
\begin{equation}
    P(x) = \frac{\exp\left(\frac{z(x)}{\tau}\right)}{\sum_{x'} \exp\left(\frac{z(x')}{\tau}\right)}
\end{equation}
where $z(x)$ represents the unnormalized logit for token $x$, and $\tau > 0$ controls the degree of randomness: a lower $ \tau $ makes the output more deterministic, while a higher $\tau$ increases diversity by flattening the probability distribution.

In video generation, a similar temperature principle can be considered when using DiT models, where the temporal attention mechanism controls the relationship between generated frames. Equation \ref{eq:temp_llm} presents a direct usage of temperature in temporal attention of DiT models.
\begin{equation}
\text{Attention}(Q, K) = \text{softmax} \left( \frac{QK^\top}{\mathbf{\tau} \cdot \sqrt{d_k}} \right)
\label{eq:temp_llm}
\end{equation}


In particular, properly increased temperatures amplify non-diagonal temporal attention, allowing the DiT model to draw global information from multiple frames during generation, leading to better spatial diversity and temporal consistency. On the other hand, setting the temperature extremely high results in uniform attention across all frames, possibly generating unexpected or prompt-irrelevant content.

However, video generation requires a proper balance between cross-frame and intra-frame attention, if we directly apply the LLM-style temperature adjustment similarly to change the original attention weights
, we always fail to enhance target cross-frame dependencies suitably.
Directly applying 
$\tau$ to temporal attention causes increasing changes as the model deepens and denoising steps accumulate, which can lead to overly smooth motion, loss of visual details, and unstable video generation, as illustrated in Appendix \ref{appendix:temp_comparison}.

\begin{figure*}
    \centering
    \includegraphics[width=\linewidth]{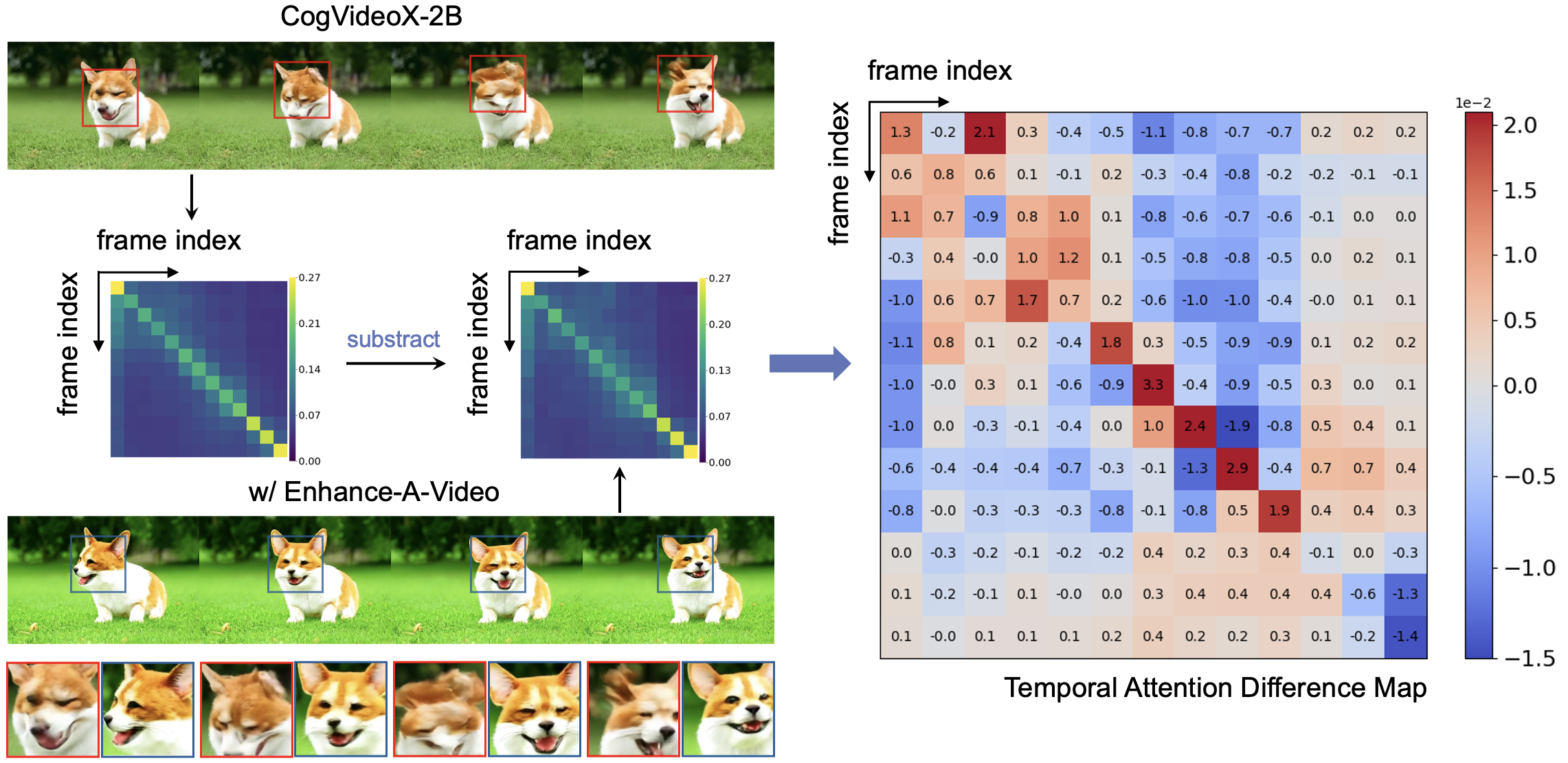}
    \caption{Temporal attention difference map between original CogVideoX model and w/ Enhance-A-Video of layer 29 at denoising step 50. Non-diagonal elements in the attention matrix of w/ Enhance-A-Video show higher values (shown in blue), while diagonal elements have reduced values (shown in red).
    }
    \label{fig:residual}
\vskip -0.15in
\end{figure*}

\subsection{Enhance Block}
To better adaptively adjust the temperature in the temporal attention mechanism, 
we propose a novel method, Enhance-A-Video, to enhance temporal consistency in video generation by utilizing the \textit{\textbf{non-diagonal temporal attention}} with \textit{\textbf{enhance temperature parameter}}. The cross-frame intensity is measured by the non-diagonal temporal attention, where higher values enable the model to focus on a broader temporal context, corresponding to higher temperature. By further introducing the enhance temperature parameter to scale the cross-frame intensity, we appropriately adjust the temporal attention outputs as a training-free enhancement.

As presented in Figure \ref{fig:enhance_block}, we design an \textbf{Enhance Block} as a parallel branch to the temporal attention mechanism. The Enhance Block operates as follows:

First, the temporal attention map $A \in \mathbb{R}^{F \times F}$ is computed, where $F$ is the number of frames. The diagonal elements $A_{ii}$ correspond to intra-frame attention, and the non-diagonal elements $A_{ij}$ ($i \neq j$) represent cross-frame attention. 

Next, the Cross-Frame Intensity (\textit{CFI}) is calculated by averaging the non-diagonal elements of the attention map:
    \begin{equation}
        \textit{CFI} = \frac{1}{F(F-1)} \sum_{i=1}^{F} \sum_{\substack{j=1 \\ j \neq i}}^{F} A_{ij}.
    \end{equation}
The \textit{CFI} is then multiplied by the enhance temperature parameter $\tau$ to enhance cross-frame correlations better:
    \begin{equation}
        \textit{CFI}_{\textit{enhanced}} = \textbf{clip} ((\tau + F) \cdot \textit{CFI}, 1).
    \end{equation}
Noticeably, the enhanced Cross-Frame Intensity ($\textit{CFI}_{\textit{enhanced}}$) is clipped at a minimum value of 1, which prevents excessive deterioration of cross-frame correlations during enhancement.

Finally, the output of the Enhance Block (\textit{$\text{CFI}_{\text{enhanced}}$}) is utilized to enhance the original temporal attention block output $\mathbf{O}_{\text{attn}}$ in the residual connection \cite{He2015DeepRL, si2023freeu}:
    \begin{equation}
    \label{eq:res}
        \mathbf{O}_{\text{final}} = \textit{CFI}_{\textit{enhanced}} \cdot \mathbf{O}_{\text{attn}} + \mathbf{H}.
    \end{equation}
where $\mathbf{H}$ represents the hidden states that are inputs of the attention block. 

When $\textit{CFI}_{\textit{enhanced}}$ exceeds 1, indicating significant cross-frame information, the ratio of temporal attention block outputs is correspondingly amplified in $\mathbf{O}_{\text{final}}$. Otherwise, the connection defaults to a standard residual connection. Since $\mathbf{O}_{\text{attn}}$ is relatively small compared to $\mathbf{H}$, modest enhancements (small $\textit{CFI}_{\textit{enhanced}}$) to $\mathbf{O}_{\text{attn}}$ slightly affect the $\mathbf{O}_{\text{final}}$ distribution, enabling Enhance-A-Video to enhance cross-frame attention without substantially altering original attention patterns. The complete analytical details are available in Appendix \ref{appendix:residual}.

The temporal attention difference map in Figure \ref{fig:residual} shows the difference between the temporal attention of the original CogVideoX model and w/ Enhance-A-Video, illustrating how Enhance-A-Video properly strengthens cross-frame attention. Specifically, certain non-diagonal elements (blue areas) are moderately increased (e.g., $0.9 \times 10^{-2}$), indicating enhanced cross-frame correlations. Meanwhile, the diagonal elements experience a minimal reduction ($3.3 \times 10^{-2}$ at most), which ensures stable intra-frame attention and preserves existing fine-grained visual details. More analysis can be found in Appendix \ref{appendix:temp_comparison}.
\begin{figure*}[ht]
    \centering
    \includegraphics[width=0.85\linewidth]{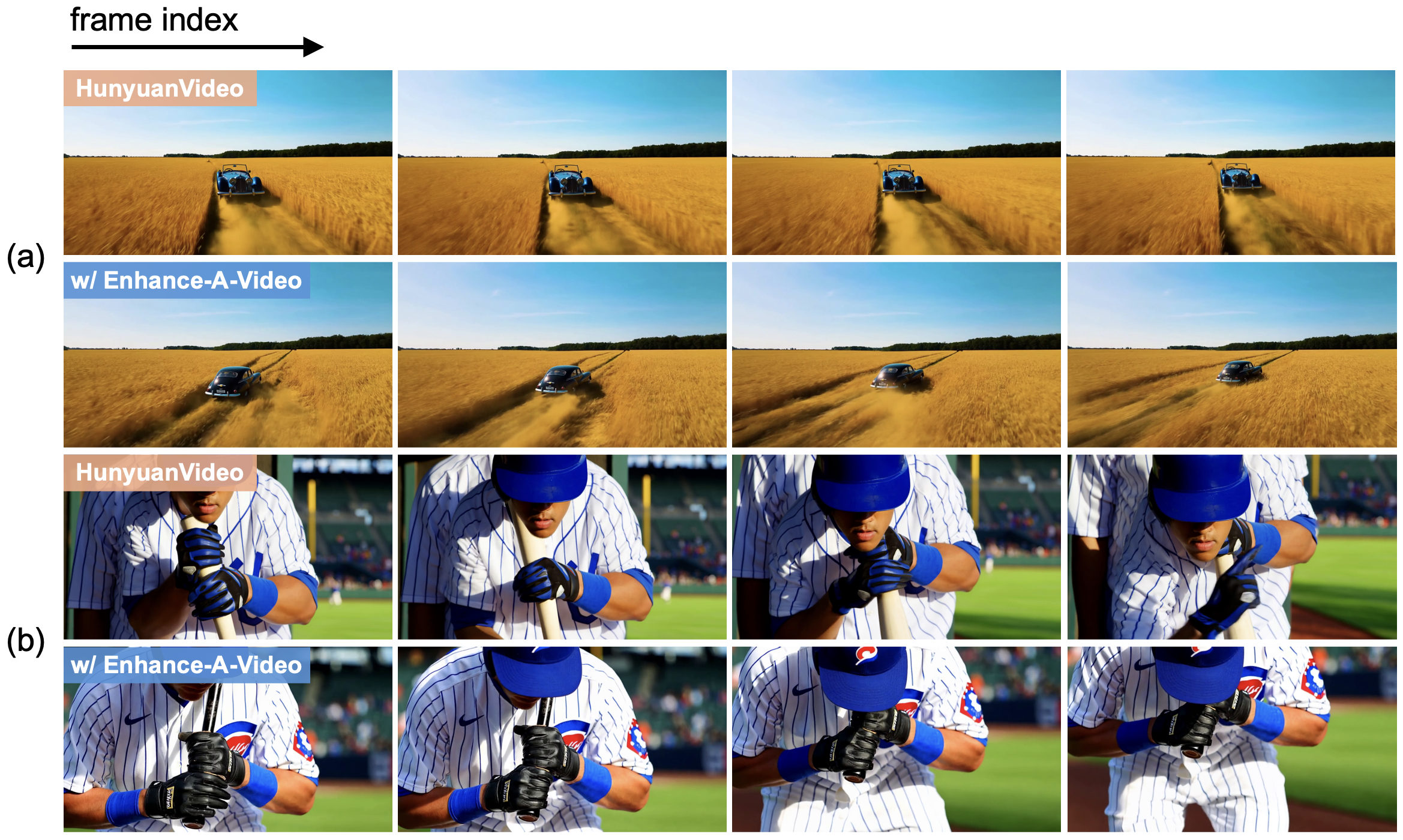}
    \caption{Qualitative results of Enhance-A-Video on HunyuanVideo. Captions: \textit{\textbf{(a)} An \textbf{antique car} drives along a dirt road through golden wheat fields. Dust rises softly as wheat brushes against the car with distant trees meeting a blue sky. \textbf{(b)} A baseball player grips a bat in \textbf{black} gloves, wearing a blue-and-white uniform and cap, with a blurred crowd and green field highlighting his focused stance.}}
    \label{fig:hunyuan}
\vskip -0.15in
\end{figure*}

\section{Experiments}
\subsection{Setup}
To evaluate the effectiveness of our proposed Enhance-A-Video method, we conduct experiments on video generation models incorporating two types of attention mechanisms: 3D full attention and spatial-temporal attention. Specifically, we choose several representative models for each category:

\textbf{3D Full Attention Model:} HunyuanVideo \cite{kong2025hunyuanvideosystematicframeworklarge}, CogVideoX \cite{yang2024cogvideoxtexttovideodiffusionmodels} and LTX-Video \cite{HaCohen2024LTXVideo}, which employ 3D full attention to model spatial and temporal dependencies simultaneously.

\textbf{Spatial-Temporal Attention Model:} Open-Sora \cite{zheng2024opensorademocratizingefficientvideo} and Open-Sora-Plan v1.0.0 \cite{lin2024opensoraplanopensourcelarge}, which decompose the attention mechanism into separate spatial and temporal components for computational efficiency and scalability.

We follow the original setup of these methods exactly and incorporate the Enhance Block exclusively into the temporal attention modules of these models during the inference phase without additional retraining or fine-tuning. For 3D full attention models, we reshape the 3D attention to focus on calculating temporal attention and the corresponding \textit{$CFI_{enhanced}$}, which is then applied to enhance the 3D attention outputs in the same way.

\subsection{3D Full Attention Model}
\begin{figure}[h]
    \centering
    \includegraphics[width=\linewidth]{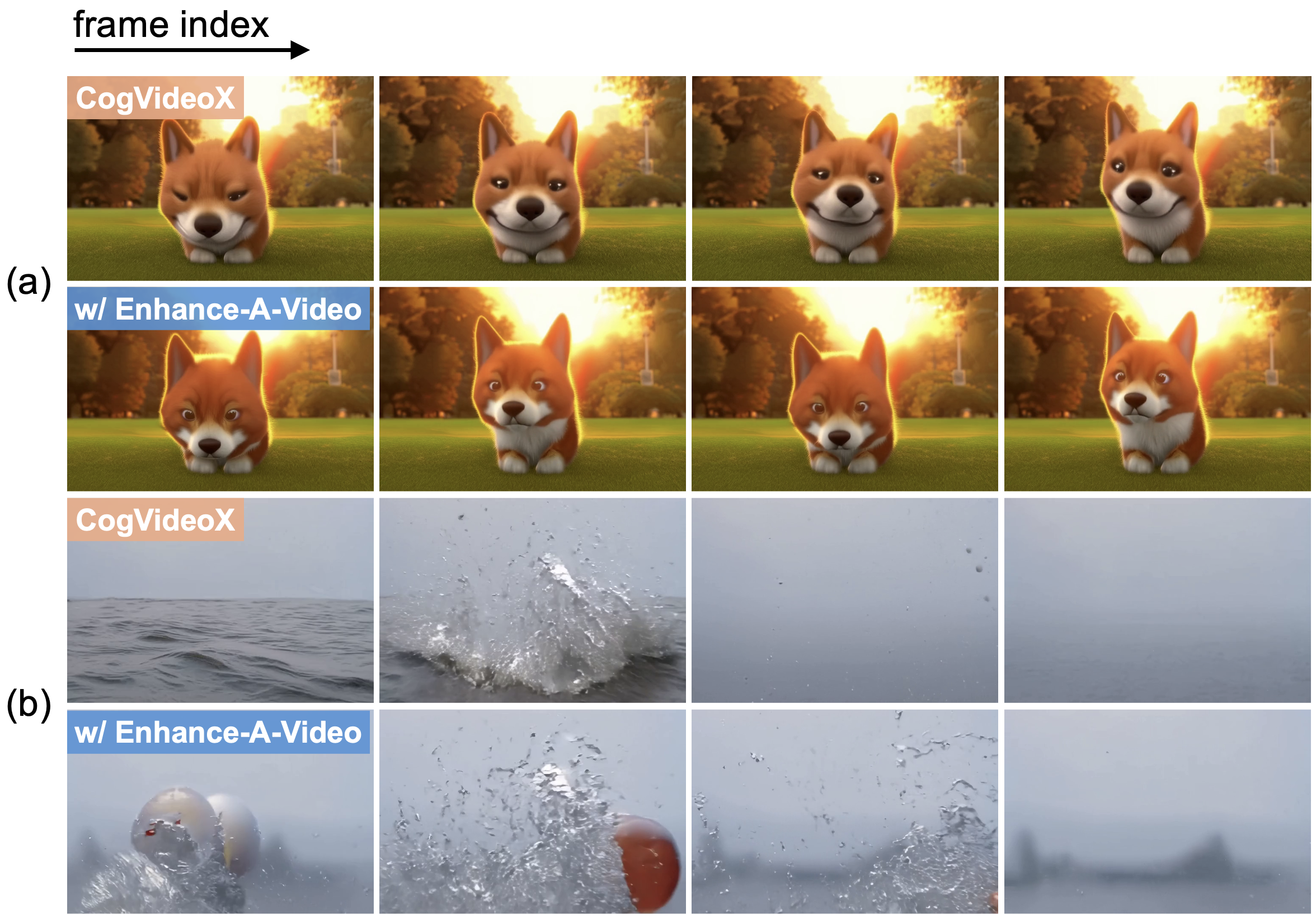}
    \caption{Qualitative results of Enhance-A-Video on CogVideoX. Captions: \textit{\textbf{(a)} A cute and happy \textbf{Corgi} playing in the park, in a surrealistic style. \textbf{(b)} \textbf{Balloon} full of water exploding in extreme slow motion.}}
    \label{fig:cogvideox}
\vskip -0.1in
\end{figure}
HunyuanVideo \cite{kong2025hunyuanvideosystematicframeworklarge} is a state-of-the-art text-to-video diffusion model recognized for its ability to produce high-resolution and temporally coherent videos from textual prompts. 
Our implementation of Enhance-A-Video augmentation in HunyuanVideo improved the model's video generation capabilities effectively. The results shown in Figure \ref{fig:hunyuan} demonstrate that Enhance-A-Video consistently produces more realistic images with better details. 

In the first case, HunyuanVideo's output shows a driverless car moving unnaturally in reverse, while Enhance-A-Video generates a car moving realistically in the correct direction. In the second case, HunyuanVideo produces conflicting artifacts - duplicate right hands and unnatural head movement. In contrast, Enhance-A-Video captures the baseball player's motion with natural fluidity and richer detail.

By applying Enhance-A-Video to CogVideoX \cite{yang2024cogvideoxtexttovideodiffusionmodels}, we observe significant improvements in prompt-video consistency, temporal coherence, and visual detail. In caption (b) of Figure \ref{fig:cogvideox}, CogVideoX fails to accurately capture the prompt describing a \enquote{balloon full of water}, generating only vague water splashes without the balloon. In contrast, the enhanced model produces videos that better align with the given prompts while delivering smoother transitions and clearer visuals.

\begin{figure}[h]
    \centering
    \includegraphics[width=0.9\linewidth]{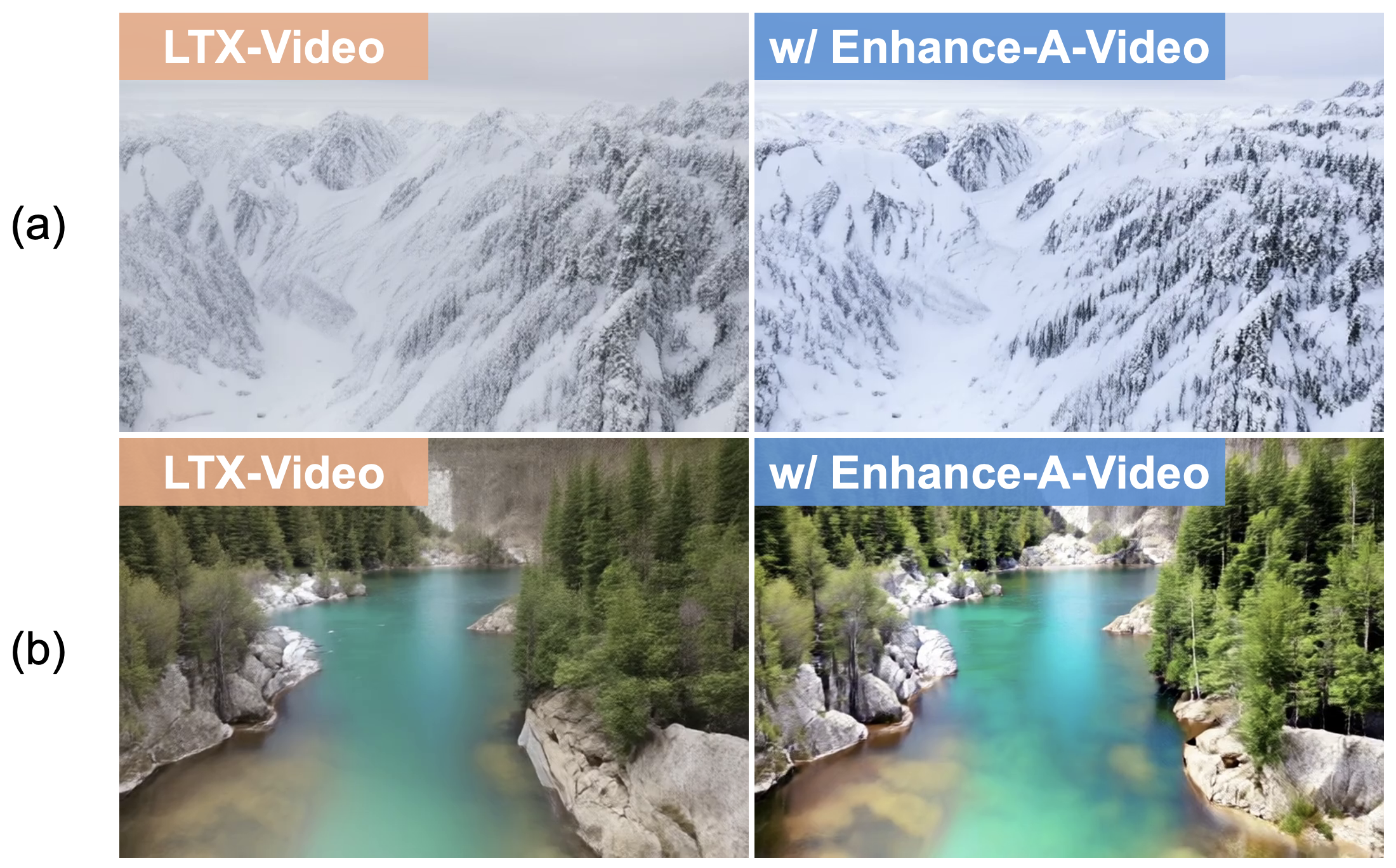}
    \vskip -0.1in
    \caption{Qualitative results of Enhance-A-Video on LTX-Video. Captions: \textit{\textbf{(a)} The camera pans over snow-covered mountains, revealing jagged peaks and deep, narrow valleys. \textbf{(b)} An emerald-green river winds through a rocky canyon, forming reflective pools amid pine trees and brown-gray rocks.}}
    \label{fig:ltxvideo}
\vskip -0.15in
\end{figure}

LTX-Video \cite{HaCohen2024LTXVideo} is a real-time latent text-to-video diffusion model that generates high-quality, temporally consistent videos efficiently. The integration of Enhance-A-Video into LTX-Video further improves temporal consistency and enhances spatial details. As exhibited in Figure \ref{fig:ltxvideo}, the enhanced model produces videos with sharper textures, more vivid colors, and smoother transitions compared to the baseline LTX-Video. 

The snow-covered mountains (top row) and river scene (bottom row) generated by Enhance-A-Video display clearer structures and more natural color gradients, while the baseline results appear less detailed and slightly blurred. This demonstrates that Enhance-A-Video effectively strengthens cross-frame attention, leading to more realistic and visually appealing videos.

\subsection{Spatial-Temporal Attention Model}
\begin{figure}[h]
    \centering
    \includegraphics[width=0.8\linewidth]{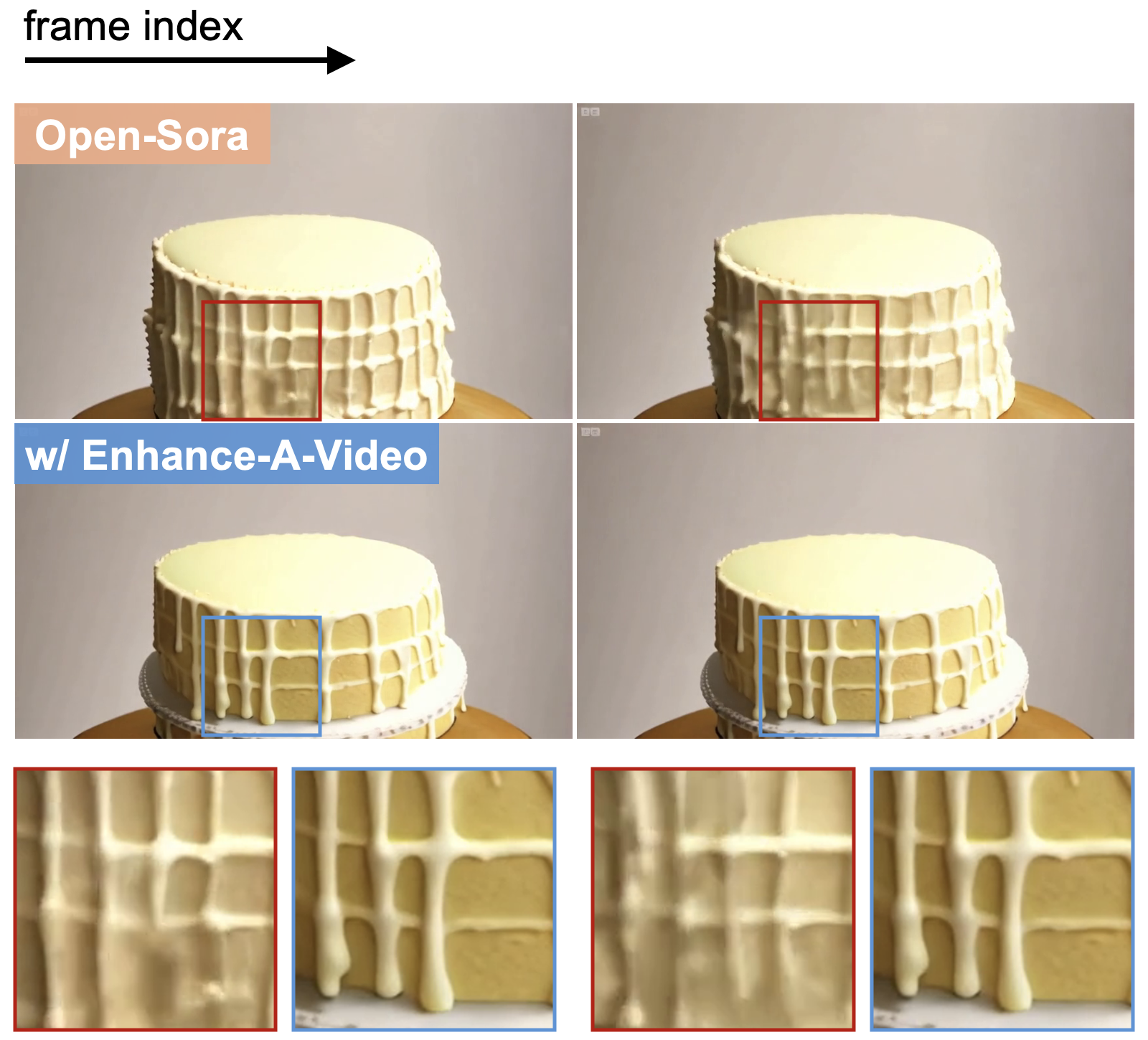}
    \vskip -0.1in
    \caption{Qualitative results of Enhance-A-Video on Open-Sora. Caption: \textit{A cake.}}
    \label{fig:opensora}
\end{figure}

Open-Sora \cite{zheng2024opensorademocratizingefficientvideo} is an efficient text-to-video generation model that utilizes a decomposed spatial-temporal attention mechanism to balance computational efficiency and video quality. Incorporating the Enhance-A-Video augmentation into Open-Sora significantly improved temporal consistency and spatial detail preservation. As demonstrated in Figure \ref{fig:opensora}, the enhanced model produces videos with more natural motion transitions and more realistic visual details.
Besides, the results on Open-Sora-Plan v1.0.0 are provided in Appendix \ref{appendix:more}.

\subsection{Quantitative Analysis}
We evaluated video quality through a blind user study of 110 participants. Each person compared two videos generated from the same text prompt and random seed - one from baseline models and one from w/ Enhance-A-Video. The videos were shown in random order to prevent bias. Participants chose which video they preferred based on three criteria: temporal consistency, prompt-video consistency, and overall visual quality.

\begin{figure}[t]
    \centering
    \includegraphics[width=0.95\linewidth]{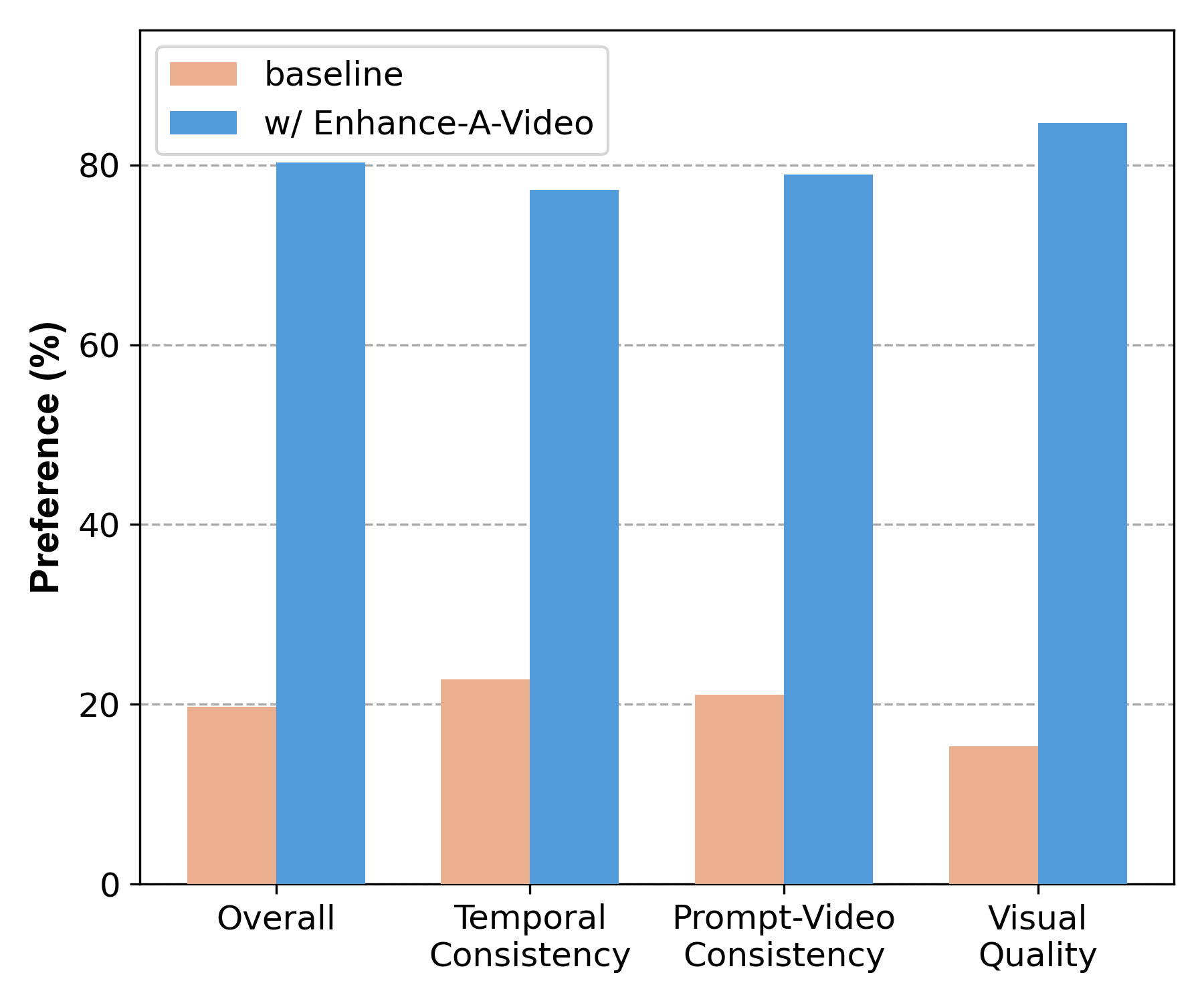}
    \vskip -0.15in
    \caption{User study results comparing baseline models and w/ Enhance-A-Video across evaluation criteria. }
    \label{fig:user_study}
\vskip -0.15in
\end{figure}

\begin{figure*}
    \centering
    \includegraphics[width=0.9\linewidth]{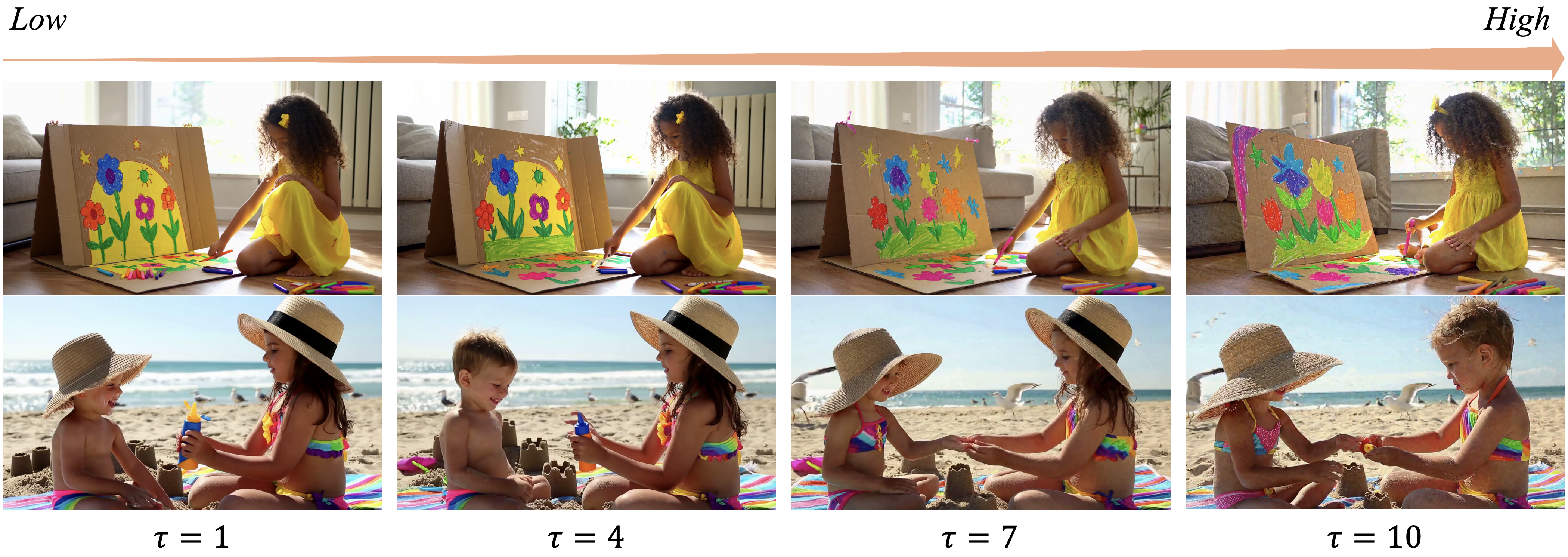}
    \vskip -0.1in
    \caption{Ablation study on the enhance temperature parameter in the Enhance Block. Moderate values balance temporal consistency and visual diversity, while extreme values degrade performance.}
    \label{fig:temp}
\vskip -0.15in
\end{figure*}

Figure \ref{fig:user_study} presents the main user study results for chosen models and w/ Enhance-A-Video of each evaluation criterion. The results show that models using Enhance-A-Video received the majority of preference, demonstrating that Enhance-A-Video notably enhances the text-to-video models' performance in all evaluated aspects \footnote{Appendix \ref{appendix:user_example} contains a comprehensive analysis with specific user study examples.}:

\textbf{Temporal Consistency.} The usage of Cross-Frame Intensity (\textit{CFI}) and the enhance temperature parameter strengthens cross-frame connections. This results in smoother motion transitions and improved frame-to-frame alignment, which creates a more stable and coherent visual experience in the generated video.

\textbf{Prompt-video Consistency.} In diffusion-based video generation, video frames are progressively denoised based on the prompt. However, the lack of temporal attention in cross-frame information transmission causes the semantic alignment between the video and the prompt to deviate gradually during generation. Enhancing cross-frame information by Enhance-A-Video ensures that objects and actions in the scene remain consistent with the prompt. This smooth semantic evolution avoids abrupt or inconsistent content, improving the alignment between the generated video and the given prompt.
 
\textbf{Visual Quality.} By using \textit{CFI} and the enhanced temperature parameter, the model makes better use of information from adjacent frames to improve details, especially in object textures and edges. The improved cross-frame attention smooths the denoising process and reduces random changes, allowing the model to generate more consistent motion and avoid unrealistic movements.

Moreover, we conducted independent evaluations using VBench \cite{huang2023vbench} for each video generation model with 5 random seeds. Table \ref{tab:vbench} shows that integrating Enhance-A-Video consistently improves VBench scores across all models \footnote{VBench may not fully reflect the quality advancements of Enhance-A-Video as discussed in Appendix \ref{appendix:discussion_vbench}.}. These results confirm that Enhance-A-Video effectively boosts temporal consistency and visual quality with minimal overhead.
\vskip -0.1in
\begin{table}[h]
    \centering
    \small
    \caption{Comparison of VBench Score for CogVideoX, Open-Sora, and LTX-Video models without and with Enhance-A-Video.}
    \vskip 0.1in
    \begin{adjustbox}{max width=\linewidth} 
    \renewcommand{\arraystretch}{1.3} 
    \begin{tabular}{lccc}
        Method & CogVideoX & Open-Sora & LTX-Video \\
        \specialrule{1pt}{1.0pt}{1.0pt} 
    Baseline & 77.27 & 79.04 & 71.93 \\
  w/ Enhance-A-Video & \cellcolor{highlight}\textbf{77.34} & \cellcolor{highlight}\textbf{79.16} & \cellcolor{highlight}\textbf{72.04} \\

    \end{tabular}
    \end{adjustbox}
    \label{tab:vbench}
    \vskip -0.1in
\end{table}
\begin{figure}
    \centering
    \includegraphics[width=0.95\linewidth]{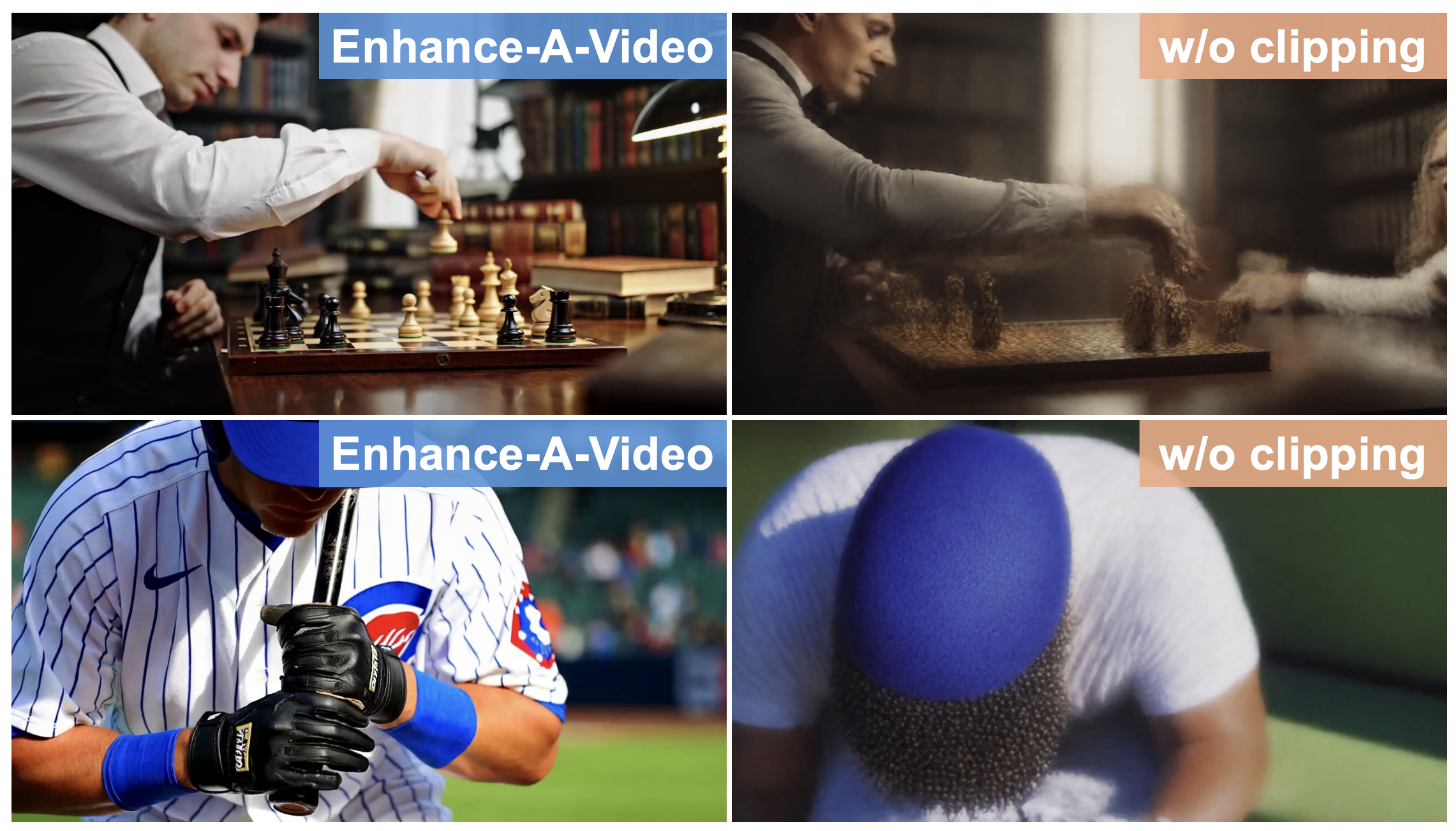}
    \vskip -0.1in
    \caption{Visual comparison of video generation results with and without the clipping mechanism in the Enhance Block.}
    \label{fig:clipping}
    \vskip -0.15in
\end{figure}
\subsection{Ablation Study}
\textbf{Impact of Temperature.} To better understand the impact of the temperature parameter, we conduct an ablation study by varying the enhance temperature parameter in the Enhance Block. Results in Figure~\ref{fig:temp} indicate that moderate temperature values achieve the best balance between temporal consistency and diversity, while extreme values (too low or too high) will degrade performance.

\textbf{Effects of Clipping.} 
Figure \ref{fig:clipping} illustrates that applying the clipping effectively stabilizes cross-frame attention, resulting in clearer visuals and smoother motion. Without clipping, the model produces noticeable artifacts such as motion blur and distorted details, highlighting the necessity of clipping for maintaining temporal consistency and preserving spatial fidelity.

\textbf{Minimal Overhead.}
To evaluate the inference efficiency of the proposed Enhance-A-Video (EAV) method, we conducted an ablation study on two prevail video generation models in Table \ref{tab:efficiency} using 1 A100 GPU. These negligible increases in the two models indicate that the Enhance-A-Video method is highly efficient and scales well when integrated into large video generation models. 
\vskip -0.2in
\begin{table}[h]
    \centering
    \small
    \caption{Comparison of inference efficiency for HunyuanVideo and CogVideoX models with and without Enhance-A-Video.}
    \vskip 0.05in
    \renewcommand{\arraystretch}{1.3} 
    \begin{tabular}{lccc}
        \multirow{2}{*}{\vspace{-0.5em}Model} & \multicolumn{2}{c}{Time (min)} & \multirow{2}{*}{\vspace{-0.5em}Overhead} \\ 
        \cmidrule(lr){2-3}
        & w/o EAV & w/ EAV & \\ 
        \specialrule{1pt}{0.5pt}{0.5pt} 
        HunyuanVideo & 50.32 & 50.72 & \cellcolor{highlight}\textbf{0.8\%} \\ 
        CogVideoX & 1.53 & 1.57 & \cellcolor{highlight}\textbf{2.1\%} \\ 
    \end{tabular}
    \label{tab:efficiency}
\end{table}
\section{Conclusion}
This paper presents Enhance-A-Video, a simple yet effective method that improves temporal consistency and visual quality in DiT-based video generation. By pioneering the exploration of cross-frame information and the temperature concept in DiT blocks, the method offers a straightforward yet powerful solution for video generation enhancement. Its robust generalization and ease of implementation suggest promising future developments in better video generation.




\section*{Impact Statement}

This paper presents research aimed at advancing video generation by improving the temporal consistency and visual quality of diffusion transformer-based models through a training-free, plug-and-play approach. Our experiments use publicly available models and benchmarks, posing no risk of harmful societal consequences while contributing to applications in entertainment, education, and media production with minimal computational overhead. 


\nocite{langley00}

\bibliography{example_paper}

\begin{thebibliography}{35}
\providecommand{\natexlab}[1]{#1}
\providecommand{\url}[1]{\texttt{#1}}
\expandafter\ifx\csname urlstyle\endcsname\relax
  \providecommand{\doi}[1]{doi: #1}\else
  \providecommand{\doi}{doi: \begingroup \urlstyle{rm}\Url}\fi

\bibitem[Blattmann et~al.(2023)Blattmann, Dockhorn, Kulal, Mendelevitch, Kilian, and Lorenz]{blattmann2023stable}
Blattmann, A., Dockhorn, T., Kulal, S., Mendelevitch, D., Kilian, M., and Lorenz, D.
\newblock Stable video diffusion: Scaling latent video diffusion models to large datasets.
\newblock \emph{arXiv preprint arXiv:2311.15127}, 2023.

\bibitem[Brooks et~al.(2024)Brooks, Peebles, Holmes, DePue, Guo, Jing, Schnurr, Taylor, Luhman, Luhman, Ng, Wang, and Ramesh]{videoworldsimulators2024}
Brooks, T., Peebles, B., Holmes, C., DePue, W., Guo, Y., Jing, L., Schnurr, D., Taylor, J., Luhman, T., Luhman, E., Ng, C., Wang, R., and Ramesh, A.
\newblock Video generation models as world simulators.
\newblock 2024.

\bibitem[Chen et~al.(2024)Chen, Xu, Ren, Cong, He, Xie, Sinha, Luo, Xiang, and Perez-Rua]{chen2024gentrondiffusiontransformersimage}
Chen, S., Xu, M., Ren, J., Cong, Y., He, S., Xie, Y., Sinha, A., Luo, P., Xiang, T., and Perez-Rua, J.-M.
\newblock Gentron: Diffusion transformers for image and video generation, 2024.
\newblock URL \url{https://arxiv.org/abs/2312.04557}.

\bibitem[Chen et~al.(2021)Chen, Kornblith, Norouzi, and Hinton]{chen2021empirical}
Chen, T., Kornblith, S., Norouzi, M., and Hinton, G.
\newblock An empirical study of training self-supervised vision transformers.
\newblock In \emph{International Conference on Computer Vision (ICCV)}, 2021.

\bibitem[Christiano et~al.(2017)Christiano, Leike, Brown, Martic, Legg, and Amodei]{christiano2017deep}
Christiano, P.~F., Leike, J., Brown, T.~B., Martic, M., Legg, S., and Amodei, D.
\newblock Deep reinforcement learning from human preferences.
\newblock \emph{Advances in neural information processing systems}, 30, 2017.

\bibitem[Esser et~al.(2024)Esser, Kulal, Blattmann, Entezari, Muller, Saini, Levi, Lorenz, Sauer, Boesel, Podell, Dockhorn, English, Lacey, Goodwin, Marek, and Rombach]{esser2024scaling}
Esser, P., Kulal, S., Blattmann, A., Entezari, R., Muller, J., Saini, H., Levi, Y., Lorenz, D., Sauer, A., Boesel, F., Podell, D., Dockhorn, T., English, Z., Lacey, K., Goodwin, A., Marek, Y., and Rombach, R.
\newblock Scaling rectified flow transformers for high-resolution image synthesis.
\newblock \emph{arXiv preprint arXiv:2403.03206}, 2024.
\newblock \doi{10.48550/arXiv.2403.03206}.

\bibitem[Gao et~al.(2024)Gao, Zhuo, Liu, Du, Luo, Qiu, Zhang, Lin, Huang, Geng, Zhang, Xi, Shao, Jiang, Yang, Ye, Tong, He, Qiao, and Li]{gao2024luminat2xtransformingtextmodality}
Gao, P., Zhuo, L., Liu, D., Du, R., Luo, X., Qiu, L., Zhang, Y., Lin, C., Huang, R., Geng, S., Zhang, R., Xi, J., Shao, W., Jiang, Z., Yang, T., Ye, W., Tong, H., He, J., Qiao, Y., and Li, H.
\newblock Lumina-t2x: Transforming text into any modality, resolution, and duration via flow-based large diffusion transformers, 2024.
\newblock URL \url{https://arxiv.org/abs/2405.05945}.

\bibitem[Guo et~al.(2024)Guo, Yang, Rao, Liang, Wang, Qiao, Agrawala, Lin, and Dai]{guo2023animatediff}
Guo, Y., Yang, C., Rao, A., Liang, Z., Wang, Y., Qiao, Y., Agrawala, M., Lin, D., and Dai, B.
\newblock Animatediff: Animate your personalized text-to-image diffusion models without specific tuning.
\newblock \emph{International Conference on Learning Representations}, 2024.

\bibitem[HaCohen et~al.(2024)HaCohen, Chiprut, Brazowski, Shalem, Moshe, Richardson, Levin, Shiran, Zabari, Gordon, Panet, Weissbuch, Kulikov, Bitterman, Melumian, and Bibi]{HaCohen2024LTXVideo}
HaCohen, Y., Chiprut, N., Brazowski, B., Shalem, D., Moshe, D., Richardson, E., Levin, E., Shiran, G., Zabari, N., Gordon, O., Panet, P., Weissbuch, S., Kulikov, V., Bitterman, Y., Melumian, Z., and Bibi, O.
\newblock Ltx-video: Realtime video latent diffusion.
\newblock \emph{arXiv preprint arXiv:2501.00103}, 2024.

\bibitem[He et~al.(2024)He, Xue, Liu, Lin, Gao, Lin, Qiao, Ouyang, and Liu]{he2024venhancergenerativespacetimeenhancement}
He, J., Xue, T., Liu, D., Lin, X., Gao, P., Lin, D., Qiao, Y., Ouyang, W., and Liu, Z.
\newblock Venhancer: Generative space-time enhancement for video generation, 2024.
\newblock URL \url{https://arxiv.org/abs/2407.07667}.

\bibitem[He et~al.(2015)He, Zhang, Ren, and Sun]{He2015DeepRL}
He, K., Zhang, X., Ren, S., and Sun, J.
\newblock Deep residual learning for image recognition.
\newblock \emph{2016 IEEE Conference on Computer Vision and Pattern Recognition (CVPR)}, pp.\  770--778, 2015.
\newblock URL \url{https://api.semanticscholar.org/CorpusID:206594692}.

\bibitem[Henschel et~al.(2024)Henschel, Khachatryan, Hayrapetyan, Poghosyan, Tadevosyan, Wang, Navasardyan, and Shi]{henschel2024streamingt2v}
Henschel, R., Khachatryan, L., Hayrapetyan, D., Poghosyan, H., Tadevosyan, V., Wang, Z., Navasardyan, S., and Shi, H.
\newblock Streamingt2v: Consistent, dynamic, and extendable long video generation from text.
\newblock \emph{arXiv preprint arXiv:2403.14773}, 2024.

\bibitem[Ho et~al.(2022)Ho, Salimans, Gritsenko, Chan, Norouzi, and Fleet]{ho2022video}
Ho, J., Salimans, T., Gritsenko, A., Chan, W., Norouzi, M., and Fleet, D.~J.
\newblock Video diffusion models.
\newblock \emph{arXiv preprint arXiv:2204.03458}, 2022.

\bibitem[Holtzman et~al.(2020)Holtzman, Buys, Du, Forbes, and Choi]{holtzman2020curious}
Holtzman, A., Buys, J., Du, L., Forbes, M., and Choi, Y.
\newblock The curious case of neural text degeneration.
\newblock In \emph{International Conference on Learning Representations (ICLR)}, 2020.

\bibitem[Huang et~al.(2024)Huang, He, Yu, Zhang, Si, Jiang, Zhang, Wu, Jin, Chanpaisit, Wang, Chen, Wang, Lin, Qiao, and Liu]{huang2023vbench}
Huang, Z., He, Y., Yu, J., Zhang, F., Si, C., Jiang, Y., Zhang, Y., Wu, T., Jin, Q., Chanpaisit, N., Wang, Y., Chen, X., Wang, L., Lin, D., Qiao, Y., and Liu, Z.
\newblock {VBench}: Comprehensive benchmark suite for video generative models.
\newblock In \emph{Proceedings of the IEEE/CVF Conference on Computer Vision and Pattern Recognition}, 2024.

\bibitem[Kong et~al.(2025)Kong, Tian, Zhang, Min, Dai, Zhou, Xiong, Li, Wu, Zhang, Wu, Lin, Yuan, Long, Wang, Wang, Li, Huang, Yang, Tan, Wang, Song, Bai, Wu, Xue, Wang, Wang, Liu, Li, Li, Wang, Yu, Deng, Li, Chen, Cui, Peng, Yu, He, Xu, Zhou, Xu, Tao, Lu, Liu, Zhou, Wang, Yang, Wang, Liu, Jiang, and Zhong]{kong2025hunyuanvideosystematicframeworklarge}
Kong, W., Tian, Q., Zhang, Z., Min, R., Dai, Z., Zhou, J., Xiong, J., Li, X., Wu, B., Zhang, J., Wu, K., Lin, Q., Yuan, J., Long, Y., Wang, A., Wang, A., Li, C., Huang, D., Yang, F., Tan, H., Wang, H., Song, J., Bai, J., Wu, J., Xue, J., Wang, J., Wang, K., Liu, M., Li, P., Li, S., Wang, W., Yu, W., Deng, X., Li, Y., Chen, Y., Cui, Y., Peng, Y., Yu, Z., He, Z., Xu, Z., Zhou, Z., Xu, Z., Tao, Y., Lu, Q., Liu, S., Zhou, D., Wang, H., Yang, Y., Wang, D., Liu, Y., Jiang, J., and Zhong, C.
\newblock Hunyuanvideo: A systematic framework for large video generative models, 2025.
\newblock URL \url{https://arxiv.org/abs/2412.03603}.

\bibitem[Langley(2000)]{langley00}
Langley, P.
\newblock Crafting papers on machine learning.
\newblock In Langley, P. (ed.), \emph{Proceedings of the 17th International Conference on Machine Learning (ICML 2000)}, pp.\  1207--1216, Stanford, CA, 2000. Morgan Kaufmann.

\bibitem[Li et~al.(2025)Li, Liu, Cao, Chen, Zhuang, Chen, He, Wang, and Qiao]{li2025diffvsrenhancingrealworldvideo}
Li, X., Liu, Y., Cao, S., Chen, Z., Zhuang, S., Chen, X., He, Y., Wang, Y., and Qiao, Y.
\newblock Diffvsr: Enhancing real-world video super-resolution with diffusion models for advanced visual quality and temporal consistency, 2025.
\newblock URL \url{https://arxiv.org/abs/2501.10110}.

\bibitem[Lin et~al.(2024)Lin, Ge, Cheng, Li, Zhu, Wang, He, Ye, Yuan, Chen, Jia, Zhang, Tang, Pang, She, Yan, Hu, Dong, Chen, Pan, Zhou, Dong, Tian, and Yuan]{lin2024opensoraplanopensourcelarge}
Lin, B., Ge, Y., Cheng, X., Li, Z., Zhu, B., Wang, S., He, X., Ye, Y., Yuan, S., Chen, L., Jia, T., Zhang, J., Tang, Z., Pang, Y., She, B., Yan, C., Hu, Z., Dong, X., Chen, L., Pan, Z., Zhou, X., Dong, S., Tian, Y., and Yuan, L.
\newblock Open-sora plan: Open-source large video generation model, 2024.
\newblock URL \url{https://arxiv.org/abs/2412.00131}.

\bibitem[Lu et~al.(2024)Lu, Liang, Zhu, and Yang]{lu2024freelongtrainingfreelongvideo}
Lu, Y., Liang, Y., Zhu, L., and Yang, Y.
\newblock Freelong: Training-free long video generation with spectralblend temporal attention, 2024.
\newblock URL \url{https://arxiv.org/abs/2407.19918}.

\bibitem[Ma et~al.(2024)Ma, Wang, Jia, Chen, Liu, Li, Chen, and Qiao]{ma2024lattelatentdiffusiontransformer}
Ma, X., Wang, Y., Jia, G., Chen, X., Liu, Z., Li, Y.-F., Chen, C., and Qiao, Y.
\newblock Latte: Latent diffusion transformer for video generation, 2024.
\newblock URL \url{https://arxiv.org/abs/2401.03048}.

\bibitem[Peebles \& Xie(2022)Peebles and Xie]{DiT}
Peebles, W.~S. and Xie, S.
\newblock Scalable diffusion models with transformers.
\newblock \emph{2023 IEEE/CVF International Conference on Computer Vision (ICCV)}, pp.\  4172--4182, 2022.
\newblock URL \url{https://api.semanticscholar.org/CorpusID:254854389}.

\bibitem[Peeperkorn et~al.(2024)Peeperkorn, Kouwenhoven, Brown, and Jordanous]{peeperkorn2024temperature}
Peeperkorn, M., Kouwenhoven, T., Brown, D.~G., and Jordanous, A.~K.
\newblock Is temperature the creativity parameter of large language models?
\newblock \emph{ArXiv}, 2024.
\newblock \doi{10.48550/arXiv.2405.00492}.

\bibitem[Renze \& Guven(2024)Renze and Guven]{Renze2024TheEO}
Renze, M. and Guven, E.
\newblock The effect of sampling temperature on problem solving in large language models.
\newblock \emph{ArXiv}, abs/2402.05201, 2024.
\newblock URL \url{https://api.semanticscholar.org/CorpusID:267547769}.

\bibitem[Si et~al.(2024)Si, Huang, Jiang, and Liu]{si2023freeu}
Si, C., Huang, Z., Jiang, Y., and Liu, Z.
\newblock Freeu: Free lunch in diffusion u-net.
\newblock In \emph{CVPR}, 2024.

\bibitem[Tan et~al.(2023)Tan, Gao, Wu, Xu, Xia, Li, and Li]{tan2023temporal}
Tan, C., Gao, Z., Wu, L., Xu, Y., Xia, J., Li, S., and Li, S.~Z.
\newblock Temporal attention unit: Towards efficient spatiotemporal predictive learning.
\newblock In \emph{Proceedings of the IEEE/CVF Conference on Computer Vision and Pattern Recognition (CVPR)}, pp.\  18770--18782, 2023.

\bibitem[Vaswani et~al.(2023)Vaswani, Shazeer, Parmar, Uszkoreit, Jones, Gomez, Kaiser, and Polosukhin]{vaswani2023attentionneed}
Vaswani, A., Shazeer, N., Parmar, N., Uszkoreit, J., Jones, L., Gomez, A.~N., Kaiser, L., and Polosukhin, I.
\newblock Attention is all you need, 2023.
\newblock URL \url{https://arxiv.org/abs/1706.03762}.

\bibitem[Xu et~al.(2024{\natexlab{a}})Xu, Zou, Huang, Chen, Liu, Cheng, Shi, and Huang]{xu2024easyanimatehighperformancelongvideo}
Xu, J., Zou, X., Huang, K., Chen, Y., Liu, B., Cheng, M., Shi, X., and Huang, J.
\newblock Easyanimate: A high-performance long video generation method based on transformer architecture, 2024{\natexlab{a}}.
\newblock URL \url{https://arxiv.org/abs/2405.18991}.

\bibitem[Xu et~al.(2024{\natexlab{b}})Xu, Park, Zhang, Zhou, Shechtman, Liu, Huang, and Liu]{xu2024videogigagandetailrichvideosuperresolution}
Xu, Y., Park, T., Zhang, R., Zhou, Y., Shechtman, E., Liu, F., Huang, J.-B., and Liu, D.
\newblock Videogigagan: Towards detail-rich video super-resolution, 2024{\natexlab{b}}.
\newblock URL \url{https://arxiv.org/abs/2404.12388}.

\bibitem[Yan et~al.(2023)Yan, Hafner, James, and Abbeel]{pmlr-v202-yan23b}
Yan, W., Hafner, D., James, S., and Abbeel, P.
\newblock Temporally consistent transformers for video generation.
\newblock In Krause, A., Brunskill, E., Cho, K., Engelhardt, B., Sabato, S., and Scarlett, J. (eds.), \emph{Proceedings of the 40th International Conference on Machine Learning}, volume 202 of \emph{Proceedings of Machine Learning Research}, pp.\  39062--39098. PMLR, 23--29 Jul 2023.
\newblock URL \url{https://proceedings.mlr.press/v202/yan23b.html}.

\bibitem[Yang et~al.(2024)Yang, Teng, Zheng, Ding, Huang, Xu, Yang, Hong, Zhang, Feng, Yin, Gu, Zhang, Wang, Cheng, Liu, Xu, Dong, and Tang]{yang2024cogvideoxtexttovideodiffusionmodels}
Yang, Z., Teng, J., Zheng, W., Ding, M., Huang, S., Xu, J., Yang, Y., Hong, W., Zhang, X., Feng, G., Yin, D., Gu, X., Zhang, Y., Wang, W., Cheng, Y., Liu, T., Xu, B., Dong, Y., and Tang, J.
\newblock Cogvideox: Text-to-video diffusion models with an expert transformer, 2024.
\newblock URL \url{https://arxiv.org/abs/2408.06072}.

\bibitem[Zhang et~al.(2023{\natexlab{a}})Zhang, Wu, Liu, Zhao, Ran, Gu, Gao, and Shou]{zhang2023show}
Zhang, D.~J., Wu, J.~Z., Liu, J.-W., Zhao, R., Ran, L., Gu, Y., Gao, D., and Shou, M.~Z.
\newblock Show-1: Marrying pixel and latent diffusion models for text-to-video generation.
\newblock \emph{arXiv preprint arXiv:2309.15818}, 2023{\natexlab{a}}.

\bibitem[Zhang et~al.(2023{\natexlab{b}})Zhang, Rao, and Agrawala]{zhang2023adding}
Zhang, L., Rao, A., and Agrawala, M.
\newblock Adding conditional control to text-to-image diffusion models, 2023{\natexlab{b}}.

\bibitem[Zheng et~al.(2024)Zheng, Peng, Yang, Shen, Li, Liu, Zhou, Li, and You]{zheng2024opensorademocratizingefficientvideo}
Zheng, Z., Peng, X., Yang, T., Shen, C., Li, S., Liu, H., Zhou, Y., Li, T., and You, Y.
\newblock Open-sora: Democratizing efficient video production for all, 2024.
\newblock URL \url{https://arxiv.org/abs/2412.20404}.

\bibitem[Zhou et~al.(2024)Zhou, Yang, Wang, Luo, and Loy]{zhou2024upscaleavideo}
Zhou, S., Yang, P., Wang, J., Luo, Y., and Loy, C.~C.
\newblock {Upscale-A-Video}: Temporal-consistent diffusion model for real-world video super-resolution.
\newblock In \emph{CVPR}, 2024.

\end{thebibliography}
\bibliographystyle{icml2025}

\newpage
\appendix

\onecolumn

\section{Temperature Method Comparison}
\label{appendix:temp_comparison}
\begin{figure*}[h]
    \centering
    \includegraphics[width=\linewidth]{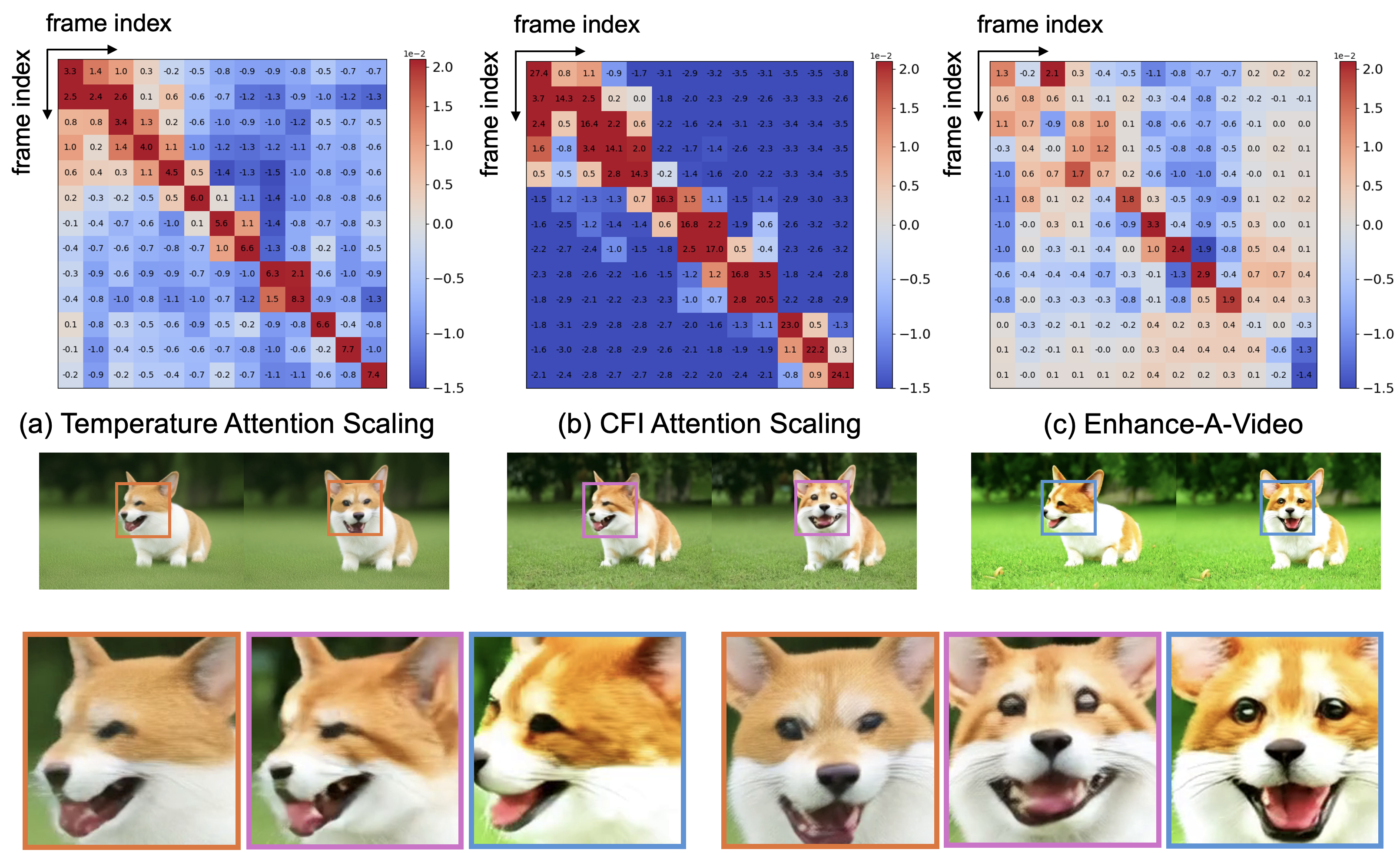}
    \caption{Temporal attention difference maps and corresponding generated videos comparing three temperature enhancement methods. (a) Temperature Attention Scaling $\tau = 1.1$. (b) CFI Attention Scaling. (c) Enhance-A-Video Method.}
    \label{fig:temp_comparison}
\end{figure*}

In Figure \ref{fig:temp_comparison}(a) and (b), where the temperature parameter $\tau$ and Cross-Frame Intensity are directly applied in temporal attention calculation separately as presented in Equation \ref{eq:temp_llm_1} and \ref{eq:temp_llm_2}, the diagonal elements (e.g., 27.4, 6.3) show a significant weakening of intra-frame attention, leading to the severe loss of spatial details and resulting in blurry and unrealistic textures. Additionally, the large negative values in the off-diagonal regions indicate an overabundant distributed enhancement of cross-frame attention, resulting in limited improvement in video quality. 
\begin{equation}
\text{Attention}(Q, K, V) = \text{softmax} \left( \frac{QK^\top}{\tau \cdot \sqrt{d_k}} \right) V
\label{eq:temp_llm_1}
\end{equation}
\begin{equation}
\text{Attention}(Q, K, V) = \text{softmax} \left( \frac{QK^\top}{\boldsymbol{\mathit{CFI}_{\mathit{enhanced}}} \cdot \sqrt{d_k}} \right) V
\label{eq:temp_llm_2}
\end{equation}
In contrast, Figure \ref{fig:temp_comparison}(c) using the Enhance-A-Video method shows modest changes along the diagonal, with values close to zero, preserving intra-frame attention and maintaining fine-grained details. Moreover, the negative values in the off-diagonal regions (e.g., -1.3, -0.9) reflect a targeted and moderate enhancement of cross-frame attention, significantly improving motion coherence and overall video quality.

\section{CFI Distribution and L2 Norm Proportion in Residual Connection}
\label{appendix:residual}
\begin{figure*}[h]
    \centering
    \includegraphics[width=0.9\linewidth]{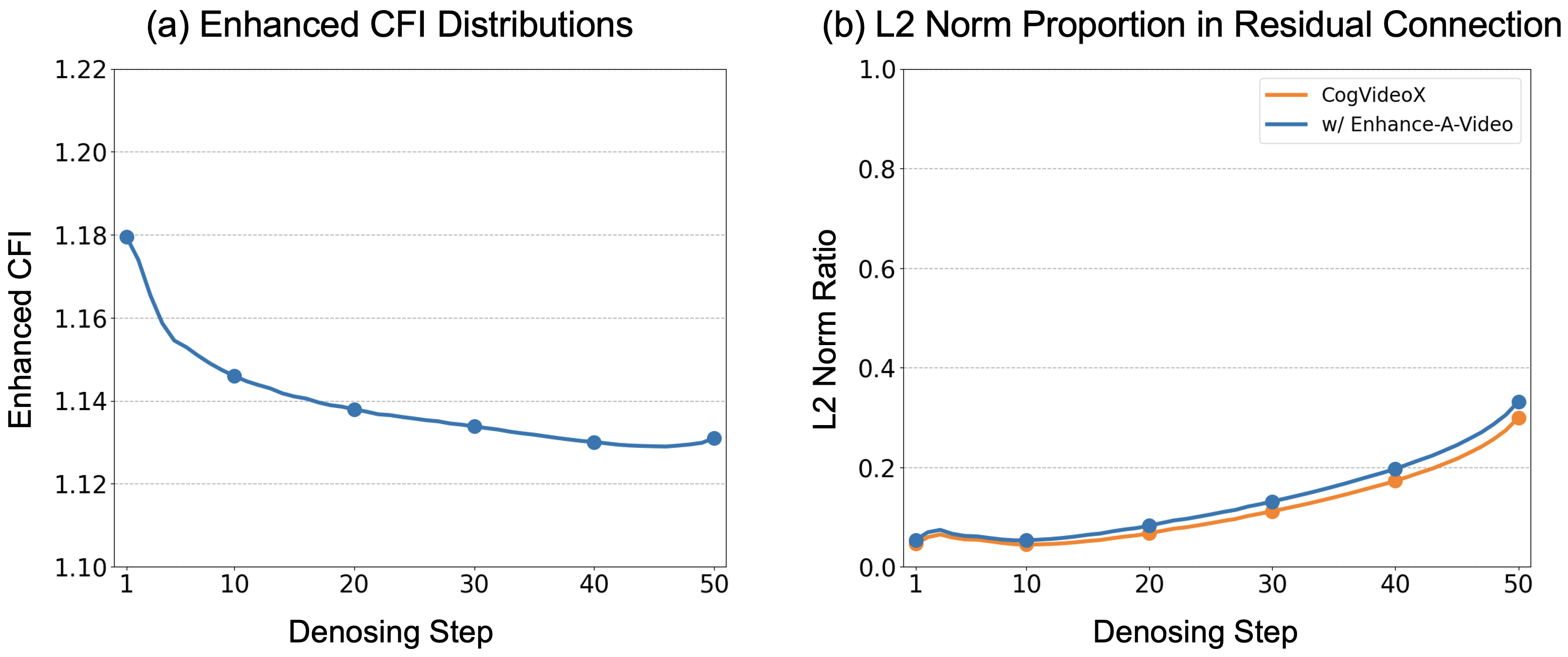}
    \caption{(a) The distribution of $\textit{CFI}_{\textit{enhanced}}$ during the inference of CogVideoX w/ Enhance-A-Video in layer 4. (b) The proportion of $l_2$ norms between $\mathbf{O}_{\text{attn}}$ and $\mathbf{H}$ in residual connection in layer 4.}
    \label{fig:res}
\end{figure*}

The $\textit{CFI}_{\textit{enhanced}}$ values in Figure \ref{fig:res}(a) range between 1.12-1.18, indicating a modest enhancement of keyframes containing important temporal information. Figure \ref{fig:res}(b) shows two low proportions calculated as follows:
\begin{equation}
    \text{prop}_{\text{CogvideoX}} = \frac{||\mathbf{O}_{\text{attn}}||_2}{||\mathbf{H}||_2}
\end{equation}
\begin{equation}
    \text{prop}_{\text{w/ Enhance-A-Video}} = \frac{||\textit{CFI}_{\textit{enhanced}} \cdot \mathbf{O}_{\text{attn}}||_2}{||\mathbf{H}||_2}
\end{equation}
suggesting that attention outputs are relatively small compared to hidden states in the residual connection. Consequently, applying $\textit{CFI}_{\textit{enhanced}}$ to attention outputs rather than attention allows for enhancing important information with minimal disruption to the original attention distribution. Thus, Enhance-A-Video improves temporal consistency while preserving existing spatial details.

\section{User Study Example}
\label{appendix:user_example}
\begin{figure*}[h]
    \centering
    \includegraphics[width=0.9\linewidth]{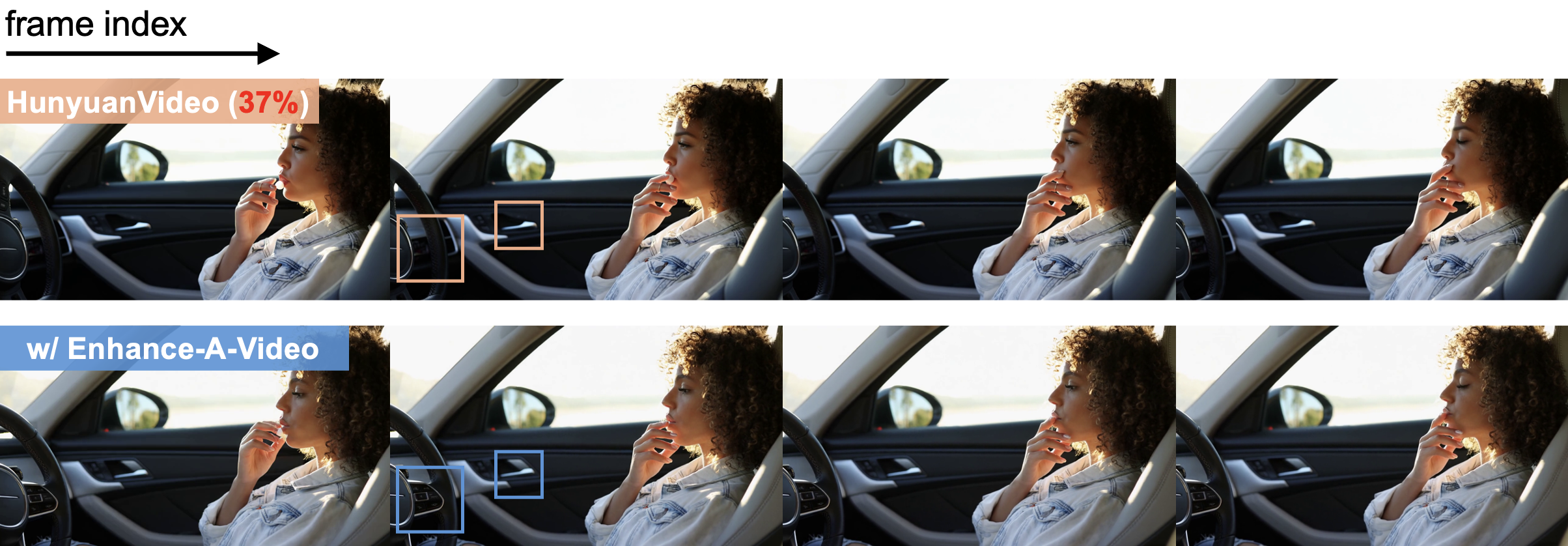}
    \caption{Selected user study example with the caption: \textit{A woman with curly hair sits comfortably in the driver's seat of a sleek, modern car, her eyes focused on the road ahead.}}
    \label{fig:user_study_bad}
\end{figure*}

\begin{figure*}[h]
    \centering
    \includegraphics[width=0.9\linewidth]{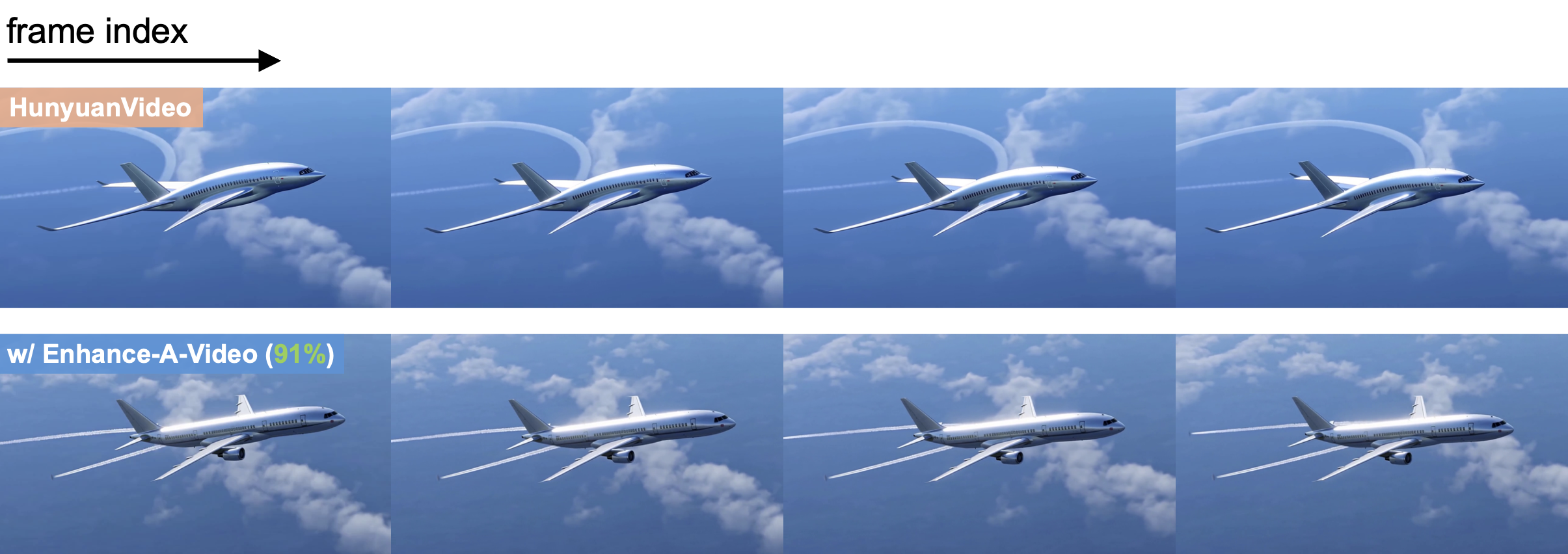}
    \caption{Selected user study example with the caption: \textit{A sleek, silver airplane soars gracefully through a vast, azure sky, its wings cutting through wispy, cotton-like clouds.}}
    \label{fig:user_study_good}
\end{figure*}

In the first example shown in Figure \ref{fig:user_study_bad}, 37\% of participants preferred the video from the basic HunyuanVideo model over the version enhanced by Enhance-A-Video. This unexpected preference occurred because these participants overlooked the enhanced details (marked by two squares) in the improved version. The enhanced model actually produced more precise and detailed elements, particularly in the interior door handle and steering wheels, demonstrating how Enhance-A-Video can improve the baseline model's ability to generate fine visual details.

Nevertheless, the enhanced result received an overwhelming 91\% of votes in the second example presented in Figure \ref{fig:user_study_good}. The superior quality of the silver plane in the enhanced version is immediately apparent, making it a much clearer improvement over the original HunyuanVideo compared to the previous example, where the differences are more subtle and require careful observation to notice.

In general, Enhance-A-Video introduces more visual details in generated videos, but the limited observation time in the user study prevented volunteers from noticing this advantage. However, with the release of advanced models like Sora, the demand for detailed and realistic video generation continues to grow. This trend underscores the growing importance of Enhance-A-Video in refining details and its role as a valuable tool for achieving higher-quality video generation.

\section{More Experimental Results}
\label{appendix:more}

\begin{figure}[!ht]
    \centering
    \includegraphics[width=0.9\linewidth]{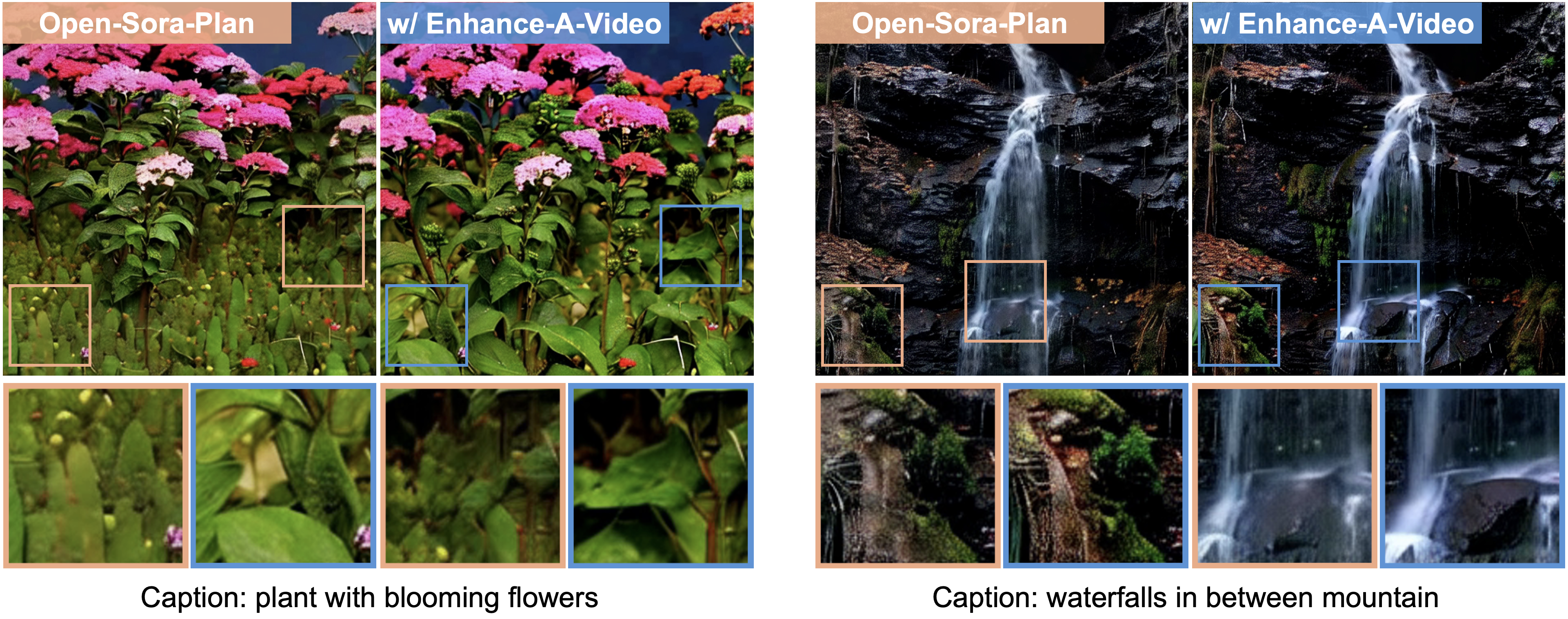}
    \caption{Qualitative results of Enhance-A-Video on Open-Sora-Plan.}
    \label{fig:opensoraplan}
\end{figure}

Open-Sora-Plan v1.0.0 \cite{lin2024opensoraplanopensourcelarge} is a text-to-video generation model leveraging a multi-resolution latent diffusion framework for high-quality and temporally coherent videos. As shown in Figure \ref{fig:opensoraplan}, for the left example, Enhance-A-Video creates clearer leaves and sharper flower details, removing the blur seen in the baseline model. In the right example, the enhanced version delivers clearer water flow and better-defined rocks, showcasing natural lighting and textures. These improvements highlight Enhance-A-Video’s ability to enhance cross-frame attention and produce visually high-quality videos.

\section{Limitations}
\label{appendix:limitation}
Our approach shows modest quantitative improvements, primarily limited by the temperature parameter requiring different optimal values for each prompt. In future work, we plan to develop an adaptive temperature mechanism using RLHF \cite{christiano2017deep} to adjust this parameter based on the specific prompt context automatically. Besides, we focused solely on enhancing temporal attention without addressing spatial attention or cross-attention mechanisms, which are crucial for preserving spatial coherence and prompt alignment. Future work could explore incorporating these mechanisms to improve spatial video quality and semantic consistency.

\section{Discussion on VBench}
\label{appendix:discussion_vbench}

\begin{figure}[!h]
    \centering
    \includegraphics[width=0.9\linewidth]{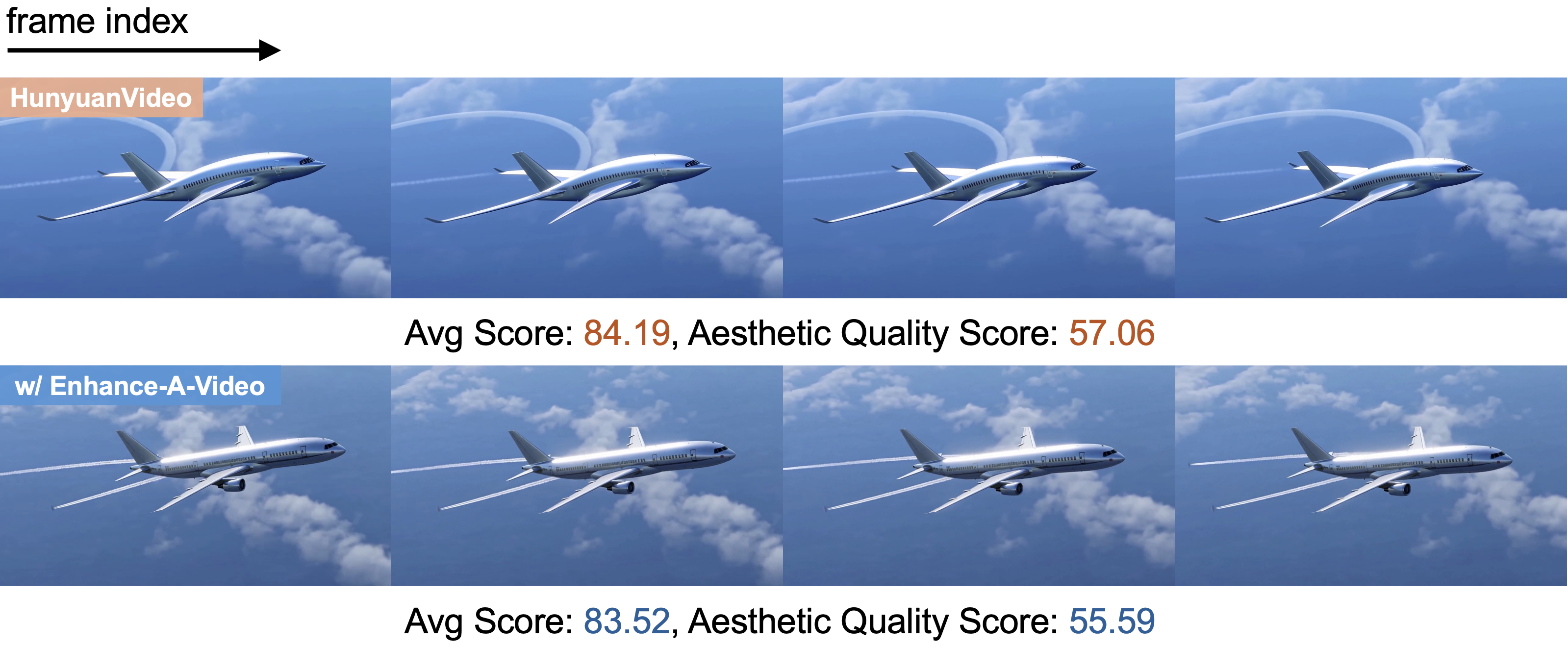}
    \caption{Comparison of video quality between HunyuanVideo and w/ Enhance-A-Video on a caption: \textit{A sleek, silver airplane soars gracefully through a vast, azure sky, its wings cutting through wispy, cotton-like clouds. The sun glints off its polished surface, creating a dazzling spectacle against the endless blue expanse. As it glides effortlessly, the contrail forms a delicate, white ribbon trailing behind, adding to the scene's ethereal beauty. The aircraft's engines emit a soft, distant hum, blending harmoniously with the serene atmosphere. Below, the earth's curvature is faintly visible, enhancing the sense of altitude and freedom. The scene captures the essence of flight, evoking a feeling of wonder and exploration.}}
    \label{fig:discussion_vbench1}
\end{figure}

The VBench benchmark does not fully reflect the substantial quality improvements achieved by Enhance-A-Video. Take the Aesthetic Quality metric as an example: the Aesthetic Quality metric in VBench is designed to evaluate the human-perceived visual quality of video frames. In the comparison of airplane footage in Figure \ref{fig:discussion_vbench1} that achieves a majority of votes from user-study participants, the Enhance-A-Video version shows noticeably better detail and clarity in rendering the aircraft compared to the HunyuanVideo baseline, yet it receives a lower Aesthetic Quality score (55.59 vs 57.06). This scoring discrepancy suggests that VBench may not effectively catch actual improvements in video enhancement quality in some cases.

\begin{figure}[!h]
    \centering
    \includegraphics[width=0.9\linewidth]{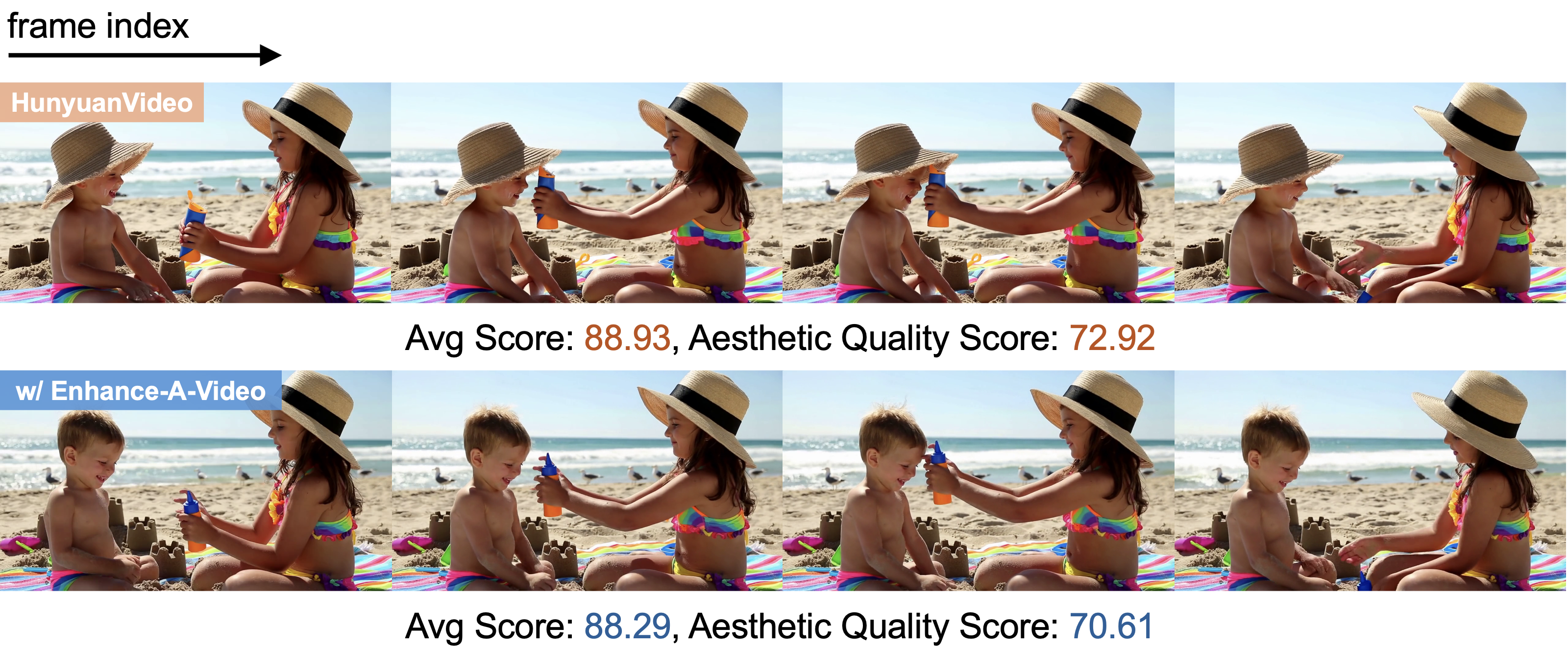}
    \caption{Comparison of video quality between HunyuanVideo and w/ Enhance-A-Video on a caption: \textit{A young girl, wearing a wide-brimmed straw hat and a colorful swimsuit, carefully applies sunblock to her younger brother's face on a sunlit beach. The boy, with sandy hair and a playful grin, sits patiently on a striped beach towel, surrounded by \textbf{sandcastles} and \textbf{beach toys}. The gentle waves of the ocean provide a soothing soundtrack as seagulls call in the distance. The girl's hands move with care, ensuring every inch of his face is protected, while the sun casts a warm glow over the scene, highlighting the siblings' bond and the carefree joy of a summer day by the sea.}}
    \label{fig:discussion_vbench2}
\end{figure}

In another example from Figure \ref{fig:discussion_vbench2}, while the video produced by Enhance-A-Video more accurately captures the prompt's details—such as \enquote{sandy hair}, \enquote{sandcastles and beach toys}—it nonetheless receives a lower Aesthetic Quality rating when compared to the baseline HunyuanVideo model.

\newpage
\section{Captions for Figure 1}
\label{appendix:abstract}
Caption 1 (top row): A young girl with curly hair, wearing a bright yellow dress, sits cross-legged on a wooden floor, surrounded by an array of colorful markers and crayons. She carefully colors a large piece of cardboard, her face a picture of concentration and creativity. The cardboard, propped up against a cozy living room couch, is filled with whimsical drawings of flowers, stars, and animals. Sunlight streams through a nearby window, casting a warm glow over her workspace. Her small hands move deftly, adding vibrant hues to her imaginative artwork, while her expression reflects pure joy and artistic focus.

Caption 2 (bottom row): A young girl, wearing a wide-brimmed straw hat and a colorful swimsuit, carefully applies sunblock to her younger brother's face on a sunlit beach. The boy, with \textbf{sandy hair} and a playful grin, sits patiently on a striped beach towel, surrounded by \textbf{sandcastles} and \textbf{beach toys}. The gentle waves of the ocean provide a soothing soundtrack as seagulls call in the distance. The girl's hands move with care, ensuring every inch of his face is protected, while the sun casts a warm glow over the scene, highlighting the siblings' bond and the carefree joy of a summer day by the sea.


\end{document}